\documentclass[10pt,twocolumn,letterpaper]{article}

\usepackage[pagenumbers]{iccv} % To force page numbers, e.g. for an arXiv version

\definecolor{iccvblue}{rgb}{0.21,0.49,0.74}
\usepackage[pagebackref,breaklinks,colorlinks,allcolors=iccvblue]{hyperref}
\usepackage{graphicx}
\usepackage{multirow}
\usepackage{rotating}
\usepackage{booktabs}
\usepackage{tabularray} 
\usepackage{algorithm}
\usepackage{algpseudocode}
\usepackage{caption}
\usepackage{ragged2e}
\usepackage[normalem]{ulem}
\usepackage{array}
\usepackage{amssymb} % For \checkmark
\usepackage{pifont} % For \xmark
\usepackage[accsupp]{axessibility}  % Improves PDF readability for those with disabilities.
\newcommand{\xmark}{\ding{55}}%

\def\paperID{14902} % *** Enter the Paper ID here
\def\confName{ICCV}
\def\confYear{2025}

\title{G\textsuperscript{2}D: Boosting Multimodal Learning with Gradient-Guided Distillation}

\author{Mohammed Rakib, Arunkumar Bagavathi\\
Oklahoma State University\\
Stillwater, Oklahoma, United States\\
{\tt\small \{mohammed.rakib, abagava\}@okstate.edu}
}
\begin{document}
\maketitle
\begin{abstract}
Multimodal learning aims to leverage information from diverse data modalities to achieve more comprehensive performance. However, conventional multimodal models often suffer from modality imbalance, where one or a few modalities dominate model optimization, leading to suboptimal feature representation and underutilization of weak modalities. To address this challenge, we introduce Gradient-Guided Distillation (\emph{G\textsuperscript{2}D}), a knowledge distillation framework that optimizes the multimodal model with a custom-built loss function that fuses both unimodal and multimodal objectives. G\textsuperscript{2}D further incorporates a dynamic sequential modality prioritization (SMP) technique in the learning process to ensure each modality leads the learning process, avoiding the pitfall of stronger modalities overshadowing weaker ones. We validate G\textsuperscript{2}D on multiple real-world datasets and show that G\textsuperscript{2}D amplifies the significance of weak modalities while training and outperforms state-of-the-art methods in classification and regression tasks. Our code is available \href{https://github.com/rAIson-Lab/G2D}{here}.
\end{abstract}

% not only learns from unimodal teacher models but also adapts dynamic gradient modulation to balance modalities in training.

% Extensive experiments demonstrate that G\textsuperscript{2}D effectively mitigates modality competition and enhances feature alignment, providing superior performance in both unimodal and multimodal scenarios.    
\section{Introduction}

% \arun{Summarize multiple aspects of multimodal representation learning and imbalance problem+its disadvantages}

% \arun{Current approaches in imbalance, their disadvantages, and open research gap}

% \arun{Real world problems based on unimodal teachers and multimodal students, and their importance}

% \arun{Motivation of the paper}

% \arun{Write Introduction as detailed and convincing as possible. Max 2 pages.}

%Performance comparison on the CREMA-D test set across audio, video, and multimodal tasks. \textbf{(a)} shows the accuracy of the audio encoder in different configurations: audio-only, audio in a multimodal model, audio in OGM-GE, and audio in G\textsuperscript{2}D. \textbf{(b)} presents the accuracy of the video encoder in similar configurations. \textbf{(c)} highlights the overall multimodal performance of baseline models, OGM-GE, and G\textsuperscript{2}D. G\textsuperscript{2}D demonstrates superior performance across all tasks, effectively addressing modality competition and achieving balanced multimodal learning.

\begin{figure*}[ht]
    \centering
    % \begin{subfigure}[t]{0.3\textwidth}
    \begin{subfigure}[t]{0.27\linewidth}
        \centering
        \includegraphics[width=\textwidth]{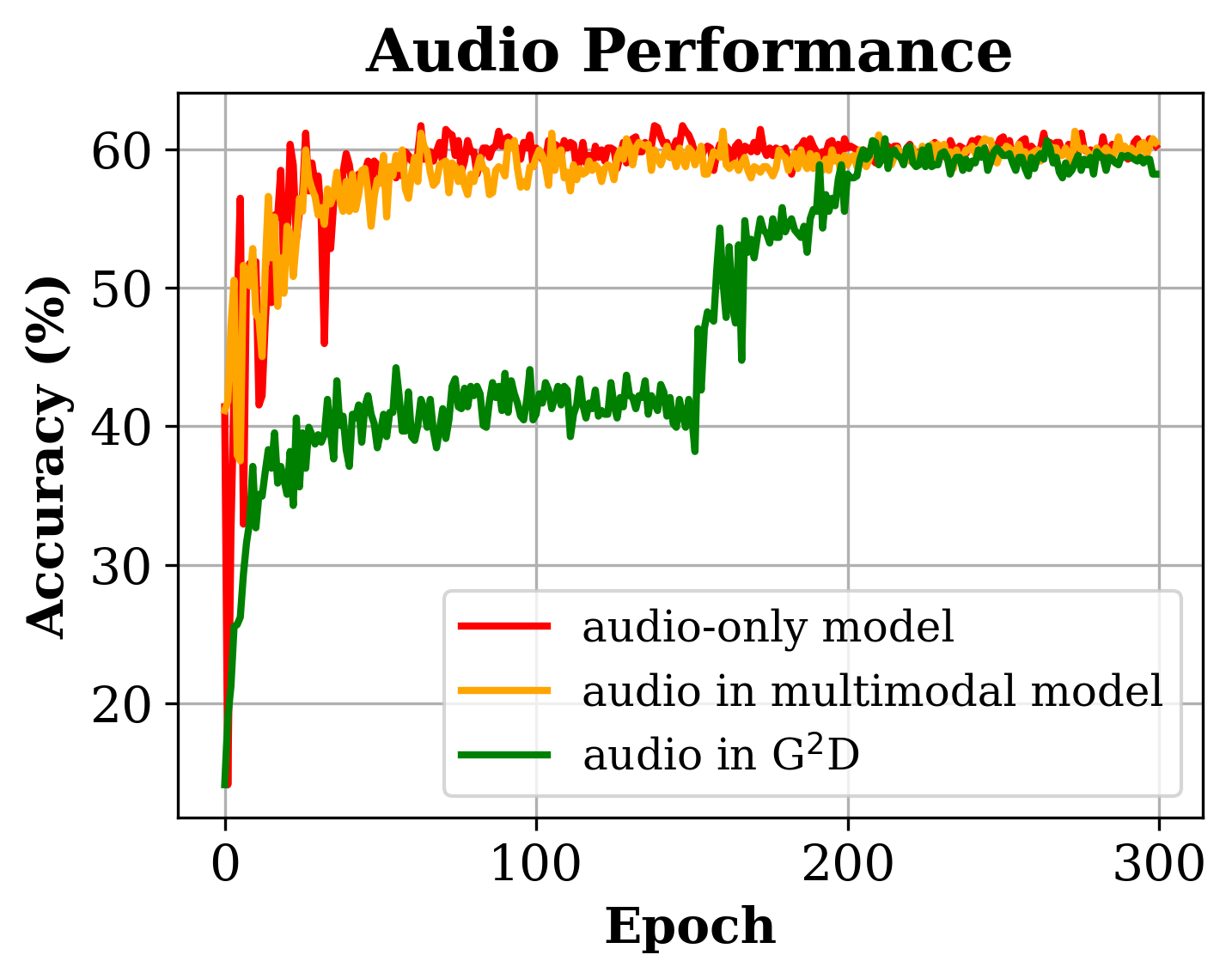}  % Update with the path to your audio plot
        \caption{Audio Performance}
        \label{fig:audio-performance}
    \end{subfigure}
    \hfill
    % \begin{subfigure}[t]{0.3\textwidth}
    \begin{subfigure}[t]{0.27\linewidth}
        \centering
        \includegraphics[width=\textwidth]{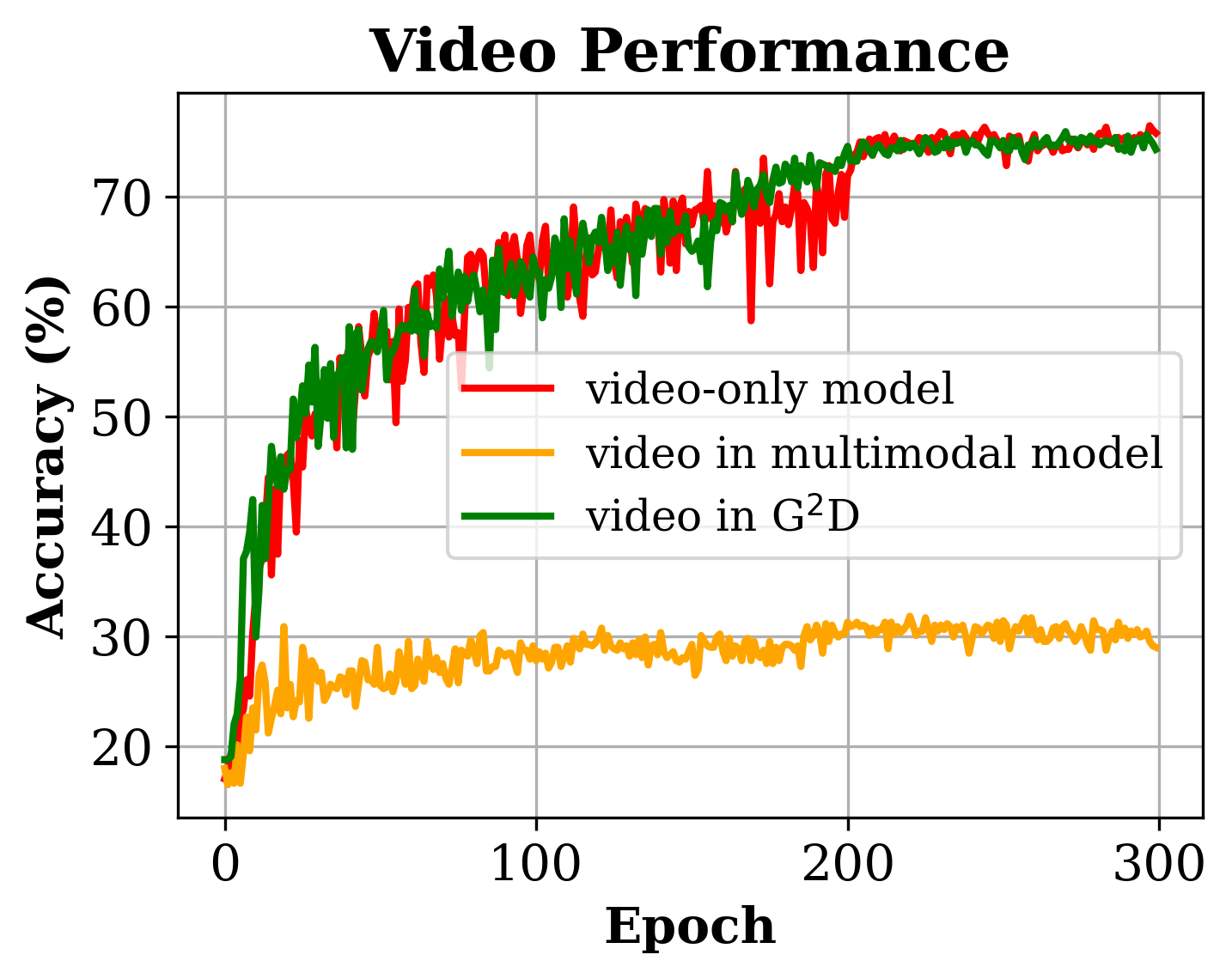}  % Update with the path to your video plot
        \caption{Video Performance}
        \label{fig:video-performance}
    \end{subfigure}
    \hfill
    % \begin{subfigure}[t]{0.3\textwidth}
    \begin{subfigure}[t]{0.27\linewidth}
        \centering
        \includegraphics[width=\textwidth]{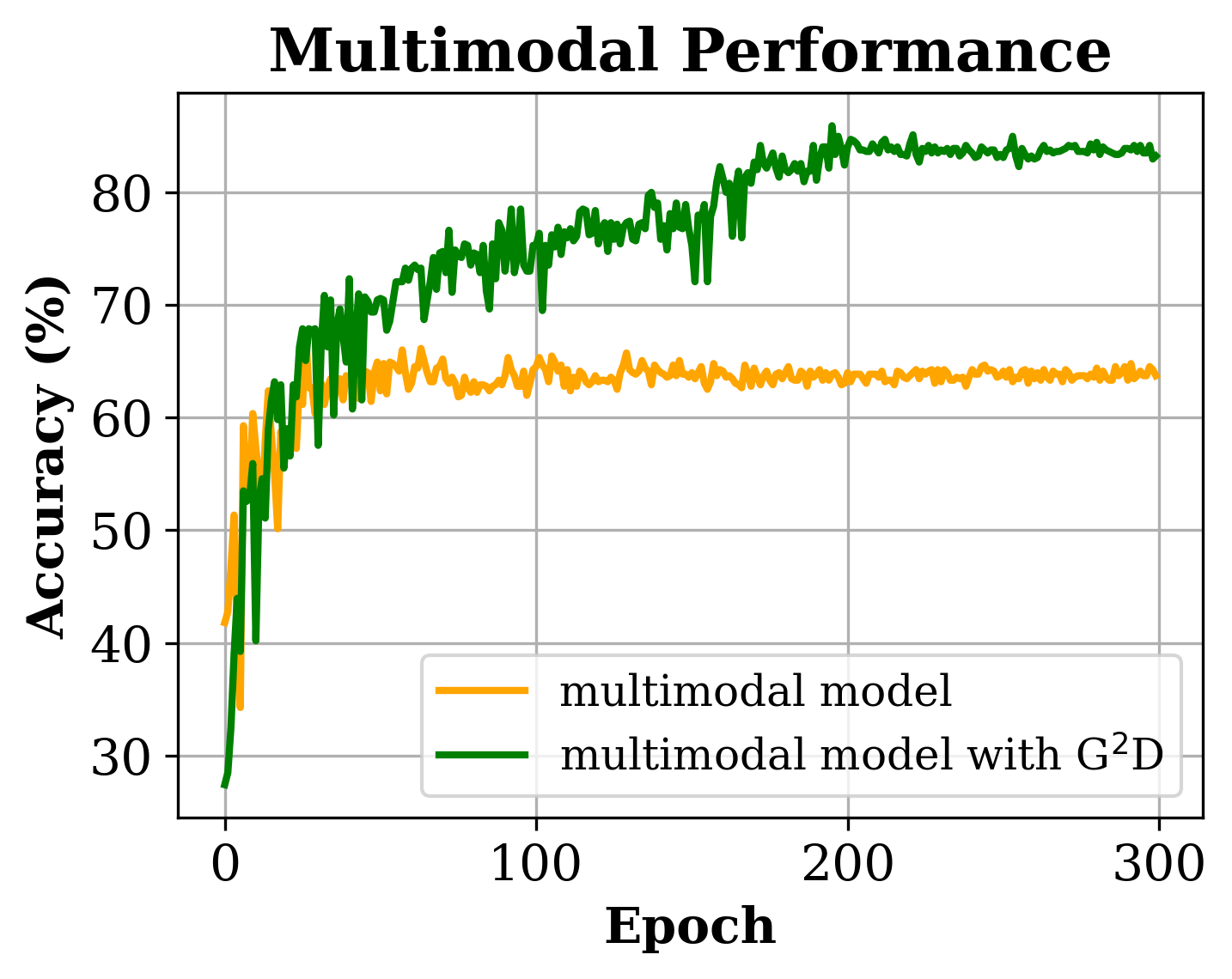}  % Update with the path to your multimodal plot
        \caption{Multimodal Performance}
        \label{fig:multimodal-performance}
    \end{subfigure}
    % \caption{Modality imbalance in the multimodal classification with CREMA-D test set. \textbf{(a)} the performance of audio modality is indifferent; \textbf{(b)} video modality is superior in unimodal setting while it is vulnerable to audio modality in a multimodal setting; \textbf{(c)} the multimodal learning performance is not optimal because of the modality imbalance. G\textsuperscript{2}D limits the optimization of superior modality and enhances the video modality to demonstrate superior multimodal performance.}
    \caption{Performance of unimodal-only, unimodal in multimodal training, and purely multimodal models on the CREMA-D test set for multimodal classification. \textbf{(a)} audio modality is indifferent to training configurations; \textbf{(b)} video modality is vulnerable to the audio modality in a multimodal setting; \textbf{(c)} performance of the multimodal model is not optimal because of modality imbalance. G\textsuperscript{2}D limits the optimization of superior modality and enhances the video modality to optimize the multimodal performance.}
    \label{fig:multimodal-imbalance}
\end{figure*}

% 

% The goal of these multimodal intelligent agents is to mimic human perception, learning, reasoning, and understanding from the basic sensory brain signals such as auditory, visual, and tactile.

% Several attempts have been made in recent years to integrate multiple data modalities in a joint learning model with an emphasis on different aspects such as multimodal alignment, fusion, representation learning, reasoning, and generation \cite{liang2024foundations,zhu2024vision+}.

Multimodal learning is one of the most prominent multidisciplinary research areas due to the increasing demand to develop intelligent agents that perceive information from diverse sensory modalities. One of the primary challenges of multimodal learning models is the \emph{modality imbalance} phenomenon~\cite{ogm-ge,agm,mm-pareto,dlmg}, also known as the modality competition~\cite{agm, modality-competition,reconboost} or modality laziness~\cite{mla,umt}. Modality imbalance occurs when one modality dominates and other modalities are underutilized in the optimization of multimodal learning models. This causes (i) inferior multimodal performance compared to unimodal models~\cite{sun2021learning,ogm-ge,g-blending}, or (ii) a larger gap in individual modality when they are optimized jointly but still improve the model performance~\cite{agm,pmr}. This imbalance occurs due to poor alignment of the modalities, model overfitting to the modalities~\cite{g-blending}, and differences in the rate of model convergence~\cite{ijcai2024p636}. An example of a modality imbalance with the multimodal dataset \emph{CREMA-D}~\cite{cremad} is given in ~\Cref{fig:multimodal-imbalance}. It is evident that the \emph{audio} features(Figure~\ref{fig:audio-performance}) dominate the video features(Figure~\ref{fig:video-performance}) when optimized in a multimodal fashion. However, video features give better performance with unimodal training. This results in sub-par performance with the joint multimodal training, shown in ~\Cref{fig:multimodal-performance}. In this work, our goal is to (i) increase the performance of weak modalities in multimodal settings and (ii) increase the overall performance of the multimodal model in downstream supervised tasks.
%Multimodal learning imbalance can result in overfitting of the model, insignificant performance in downstream tasks, and the model not being optimized to learn valuable knowledge from all available data modalities.\arun{Need references.}
%The imbalance occurs when one data modality dominates the learning model over other modalities, preventing the model from fully utilizing the valuable knowledge available in weak modalities. 

% In a different perspective, it has been studied that altering the modality gap~\cite{modalitygap} has a significant improvement in multimodal models such as CLIP~\cite{radford2021learning}.
% Traditional approaches with simple fusion architectures~\cite{g-blending,gan2020music} on multiple unimodal models will not be sufficient since each modality updates its parameters independently~\cite{ogm-ge}.

In recent years, many methods have been proposed to address modality imbalance in multimodal learning \cite{g-blending,ogm-ge,agm,greedy,mses,mslr,umt,pmr,mm-pareto,mla,dlmg,reconboost}.  \emph{Gradient modulation} is one of the popular approaches in state-of-the-art methods to dynamically modify multimodal optimization gradients and maximize equal contributions from all modalities. A common form of gradient modulation is dynamically increasing the gradients of weak modalities only for late fusion~\cite{ogm-ge,qmf,mslr} or for any type of fusion methods~\cite{agm,pmr} during the training process. Multiple variations of these gradient modulations also exist, for example, alternating gradients of each modality~\cite{mla} and controlling the dominant modality gradients~\cite{mses}. Although there exist multiple gradient modulation methods, there are very few methods that optimize both unimodal and multimodal learning objectives~\cite{mm-pareto,umt} to add the benefits of both worlds. We aim to follow this trend in this work and introduce a novel optimization strategy by incorporating knowledge distillation.

%One of the popular approaches to mitigate this problem was OGM-GE, which provided some improvements by aligning the modalities more effectively. However, as demonstrated in \cref{fig:audio-performance}, \cref{fig:video-performance}, and \cref{fig:multimodal-performance}, our proposed G\textsuperscript{2}D method significantly outperforms OGM-GE on the CREMA-D dataset. 
% Most existing methods are based on optimizing the gradients of dominant or inferior modalities for balanced learning across all modalities.
%One of the key challenges in multimodal learning is ensuring a balanced contribution from each modality while minimizing the effect of modality competition, where stronger modalities overshadow weaker ones.

In this work, we aim to utilize the full potential of both unimodal and multimodal learning for the given multimodal downstream task. We propose a framework \emph{Gradient-Guided Distillation} (\emph{G\textsuperscript{2}D}) that transfers knowledge from multiple unimodal teachers to multimodal student models. \textbf{\textit{Novelty}} of \emph{G\textsuperscript{2}D} is its use of knowledge distillation with a new learning objective and gradient modulation technique to mitigate modality imbalance and produce state-of-the-art results in multimodal learning. As depicted in Figure~\ref{fig:multimodal-imbalance}, G\textsuperscript{2}D improves the feature quality of multimodal encoder, allowing both the audio and video encoders to approach the accuracy of their unimodal counterparts when integrated into the multimodal model. This leads to an overall more balanced and better-performing multimodal model while addressing the modality imbalance issue (\Cref{fig:multimodal-performance}). Our \emph{three-fold} contributions in this work are:
% : (i) a new optimization method designed for knowledge distillation to combine contributions of unimodal and multimodal encoders, and (ii) a new dynamic sequential modality prioritization process to balance the modalities in multimodal learning.

\begin{itemize}
\item We introduce a knowledge distillation framework called G\textsuperscript{2}D that adapts a new optimization technique to fuse both unimodal and multimodal learning to enhance the performance of downstream tasks.
%  to address the multimodal imbalance problem in joint training, and also

\item We propose a new Sequential Modality Prioritization strategy that dynamically balances the optimization of weak and dominant modalities to mitigate imbalance.
%nd demonstrate that a multimodal student taught by multiple unimodal teachers is both more accurate and better calibrated than the same multimodal student trained from scratch.

\item With extensive experiments on \emph{six} real-world datasets, we show that G\textsuperscript{2}D minimizes modality imbalance and achieves superior performance. We also show that our approach is adaptive to existing methods.
%\item We show the superior performance of G\textsuperscript{2}D on supervised tasks with extensive experiments on \emph{six} real-world datasets that include a total of \emph{five} data modalities. We also show that our approach is adaptive to existing multimodal learning methods.

%The results validate the proposed G\textsuperscript{2}D approach and provide insights on how to improve multimodal students by leveraging multiple unimodal teachers.
% \item \textit{(Yet to be decided)} We demonstrate the effectiveness of MM-KD trained student models, showing that an MM-KD trained multimodal model serves as a better teacher than a regular multimodal teacher for training unimodal students.
\end{itemize}

\section{Related Work}

% processes unimodal inputs through modality-specific teacher and student encoders, generating feature representations and logits for each modality. These features are combined in the multimodal student model via a fusion module, with the output, along with teacher representations, feeding into

% \textbf{Overview of G\textsuperscript{2}D Framework.} The framework processes unimodal inputs through modality-specific teacher and student encoders, generating feature representations and logits for each modality. These features are combined in the multimodal student model via a fusion module, with the output, along with teacher representations, feeding into the $\mathcal{L}_{\text{G\textsuperscript{2}D}}$ Loss Module (Section 3.1), which consists of student loss, feature distillation loss, and logit distillation loss. Confidence scores from the \textbf{Scoring Module} (Section 3.2) are used by the \textbf{Gradient-Guided Distillation Module} (Section 3.3) to generate modulation coefficients that adaptively adjust the gradients of each modality encoder, ensuring balanced contributions across all modalities.

\begin{figure*}[!t]
\centering
\includegraphics[width=0.92\linewidth]{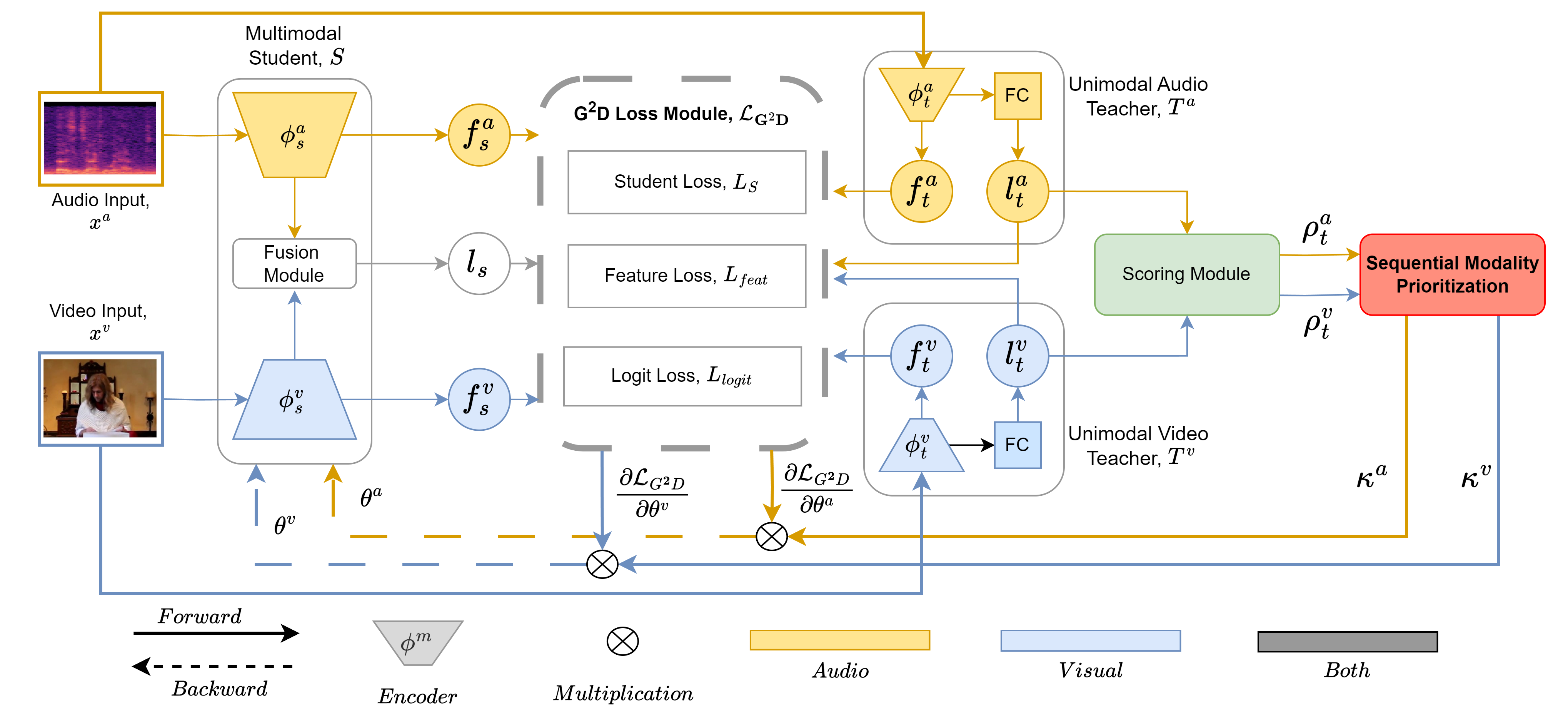}
\caption{\textbf{G\textsuperscript{2}D} consists of multiple, independently optimized \emph{unimodal teacher} encoders and jointly optimized \emph{multimodal student} encoders with all encoders generating feature representations and logits for each modality. The \textbf{$\mathcal{L}_{\text{G\textsuperscript{2}D}}$ Loss Module} consists of student loss, feature distillation loss, and logit distillation loss. Confidence scores from the \textbf{Scoring Module} are used by the \textbf{Sequential Modality Prioritization Module} to generate dynamic modulation coefficients that adaptively adjust the gradients of each encoder to ensure balanced contributions.}
\label{fig:main-diag}
\end{figure*}

%\subsection{Multimodal Knowledge Distillation}

%  such as vision-language tasks, audio-visual learning, and multimodal sentiment analysis
Knowledge distillation (KD) transfers knowledge from a larger and more complex model (\emph{teacher}) to a smaller and more efficient model (\emph{student})~\cite{hinton,modelcompression,tian-kd}. Multimodal KD extends this concept by leveraging information from multiple modalities to enhance learning, which supports several real-world AI applications (\cite{xu2021multimodal,li2020align,radevski2023multimodal,miccai2023learnable,yuan2024multiscale}). Multimodal KD distills knowledge from one modality to improve performance of other modalities or in multimodal downstream tasks. Multimodal KD has demonstrated substantial benefits, including improved performance, better multimodal alignment, and enhanced generalization across modalities (\cite{huo2024c2kd,xue2023modality,gou2021knowledge,yuan2024multiscale}). Multimodal KD has been used in multiple real-world problems like medical imaging~\cite{miccai2023learnable} to address missing modality and action recognition~\cite{radevski2023multimodal} to transfer knowledge from multimodal ensemble to a unimodal model. In this paper, we build upon recent advances in multimodal KD~\cite{umt}, specifically targeting the modality imbalance problem during multimodal training. By utilizing unimodal teacher models to guide the multimodal student model, we ensure balanced learning across modalities.

Several methods have been proposed to mitigate the modality imbalance through gradient modulation, feature rebalancing, or modality-specific learning rate adjustments (\cite{ogm-ge,agm,mses,mslr,qmf, pmr}). \emph{Gradient modulation} techniques dynamically adjust gradients to balance modality contributions during training~\cite{ogm-ge,agm}. However, these methods often require careful tuning of hyperparameters, which can limit their generalizability. \emph{Feature rebalancing} aims to optimize multimodal interaction by adjusting the contribution of each modality by enhancing the performance of unimodal learners~\cite{mla,mm-pareto,reconboost,d&r}. There is another perspective of alleviating modality imbalance with a proper label fitting using contrastive learning~\cite{dlmg}. 
% In the context of multimodal  KD, there is only a limited work that utilizes unimodal teachers as a guide for a multimodal student model (UMT)~\cite{umt-old} or simply aggregate predictions of unimodal teachers (UME)~\cite{umt}. However, this requires an empirical decision-making process, which limits its applicability across different tasks and datasets.
In the context of multimodal KD, limited work has used unimodal teachers to supervise a multimodal student. UMT~\cite{umt-old} directly supervises with unimodal teachers, while UME~\cite{umt} aggregates their logits. Choosing between them requires empirical tuning, limiting adaptability across tasks and datasets.
In this work, we propose the G\textsuperscript{2}D framework to address these limitations by combining multimodal knowledge distillation with a new gradient modulation technique. Unlike previous approaches that require careful hyperparameter tuning (\cite{ogm-ge, agm,umt, umt-old}), our approach dynamically suppresses dominant modalities based on insights from unimodal teachers. This not only makes our method applicable across different settings but also allows underrepresented modalities to learn more effectively.

% For example, by employing alternating unimodal adaptation~\cite{mla}, and by targeting gradient conflict between unimodal and multimodal losses~\cite{mm-pareto}.

\section{Methodology}

We propose \emph{Gradient-Guided Distillation} (G\textsuperscript{2}D) to mitigate modality imbalance in multimodal learning through a combination of knowledge distillation and sequential modality prioritization. The proposed G\textsuperscript{2}D adapts to labeled multimodal datasets $\mathcal{D} = {(x_i, y_i)}_{i=1}^N$,where each sample $x_i$ consists of $k$-modality inputs, denoted as $x_i = (x_i^{m_1}, x_i^{m_2}, \dots, x_i^{m_k})$, and an associated label $y_i$.

G\textsuperscript{2}D, as illustrated in Figure~\ref{fig:main-diag}, combines unimodal and multimodal learning by distilling the knowledge from unimodal teachers to the multimodal student with a new learning objective $\mathcal{L}_{\text{G\textsuperscript{2}D}}$. The \emph{scoring module} in G\textsuperscript{2}D determines modality-specific scores based on the knowledge of multiple unimodal teachers. The proposed \emph{Sequential Modality Prioritization} dynamically identifies inferior modalities and empirically modulates the gradients of multimodal student encoders to mitigate modality imbalance. 
%We will detail these three contributions in the following sections. 

%The proposed \emph{gradient-guided distillation} module dynamically identifies the inferior modalities and sequentially modulates the gradients of modality-specific student encoders in multimodal learning

%We will describe the three key stages of G\textsuperscript{2}D: \textit{G\textsuperscript{2}D loss function}, \textit{quantifying the confidence of each modality}, and \textit{modulating gradients with sequential modality prioritization} in the following sections.
%We processes unimodal inputs—such as \textit{audio} and \textit{visual}—through corresponding modality-specific teacher and student encoders. Each teacher encoder generates feature representations and logits, which serve as distillation targets for the student model, while the logits alone are used in the \textit{scoring module} to calculate confidence scores. The student encoders produce features that are fused through the \textit{fusion module} to create a multimodal representation. These fused logits are then used by the \textit{G\textsuperscript{2}D loss module}, which integrates student, feature distillation, and logit distillation losses for effective training. The confidence scores from the \textit{scoring module} are used to generate modulation coefficients in the \textit{gradient-guided distillation module}. These coefficients then adaptively modulate gradient updates for each modality-specific encoder, thereby mitigating modality imbalance.

\subsection{G\textsuperscript{2}D Loss Function}

%Let $\mathcal{D} = {(x_i, y_i)}_{i=1}^N$ represent the training dataset, where each sample $x_i$ consists of inputs from $k$ different modalities, denoted as $x_i = (x_i^{m_1}, x_i^{m_2}, \dots, x_i^{m_k})$. The corresponding ground truth label for each sample is $y_i \in {1, 2, \dots, C}$, where $C$ is the number of classes.

%We follow the traditional objective function of knowledge distillation: $\mathcal{L}_{KD} = \mathcal{L}_S + \mathcal{L}_{feat} + \mathcal{L}_{logit}$ in this work, where 

% These feature representations are subsequently passed through a linear classifier $\psi_t^m$, resulting in logits $l_t^m = \psi_t^m(f_t^m)$.
We extend traditional multimodal knowledge distillation \cite{umt} by introducing a new training objective ($\mathcal{L}_{\text{G\textsuperscript{2}D}}$) that combines \emph{unimodal feature distillation} ($\mathcal{L}_{feat}$) loss and \emph{unimodal logit distillation} ($\mathcal{L}_{\text{logit}}$) loss with the \emph{multimodal student loss} ($\mathcal{L}_S$) to mitigate modality imbalance. 
We define the unimodal teacher model $\{T^{m}\}_{m=1}^k$ and the multimodal student model $[S]_1^k$. Each teacher model $T^m$ is responsible for a single modality $m$ and consists of an encoder $\phi_t^m$ parameterized by $\theta_t^m$, which produces corresponding feature representations $f_t^m = \phi_t^m(x_i^m; \theta_t^m)$. We represent logits $l_t^m$ of teacher models after passing $f_t^m$ through a linear classifier $\psi_t^m$. Similarly, the student model $[S]_1^k$ processes the multimodal input $x_i$ through modality-specific encoders $\phi_s^m$, parameterized by $\theta_s^m$, to obtain student feature representations $f_s^m = \phi_s^m(x_i^m; \theta_s^m)$. These features are then combined through a traditional multimodal fusion module $\Phi_{\text{fusion}}(f_s^{m_1}, f_s^{m_2}, \dots, f_s^{m_k})$, to produce a multimodal representation, which is used to calculate multimodal student logits $l_s = \psi_s(\Phi_{\text{fusion}}(f_s^{m_1}, f_s^{m_2}, \dots, f_s^{m_k}))$. The proposed $\mathcal{L}_{\text{G\textsuperscript{2}D}}$ comprises three key components:

% The fusion module $\Phi_{\text{fusion}}$ can represent different fusion strategies such as summation, concatenation, late fusion \cite{late-fusion}, cross-attention \cite{cross-attention}, FiLM \cite{film}, and BiLinear Gated fusion \cite{gated}.

\begin{figure*}[!t]
    \centering
    \begin{subfigure}{0.25\textwidth}
        \includegraphics[width=1.0\linewidth]{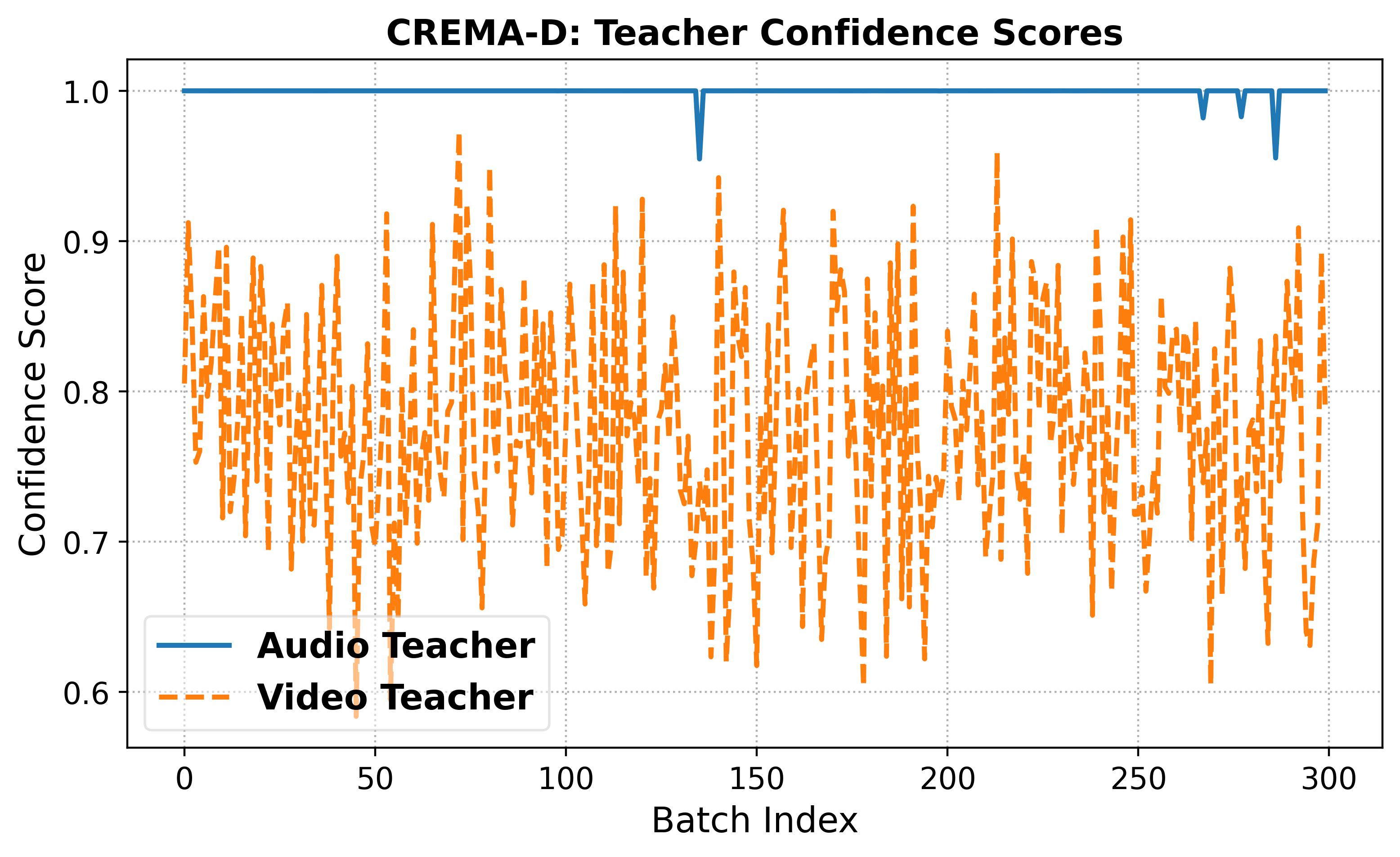}
        \caption{CREMA-D}
        \label{fig:sub1}
    \end{subfigure}%
    \begin{subfigure}{0.25\textwidth}
        \includegraphics[width=\linewidth]{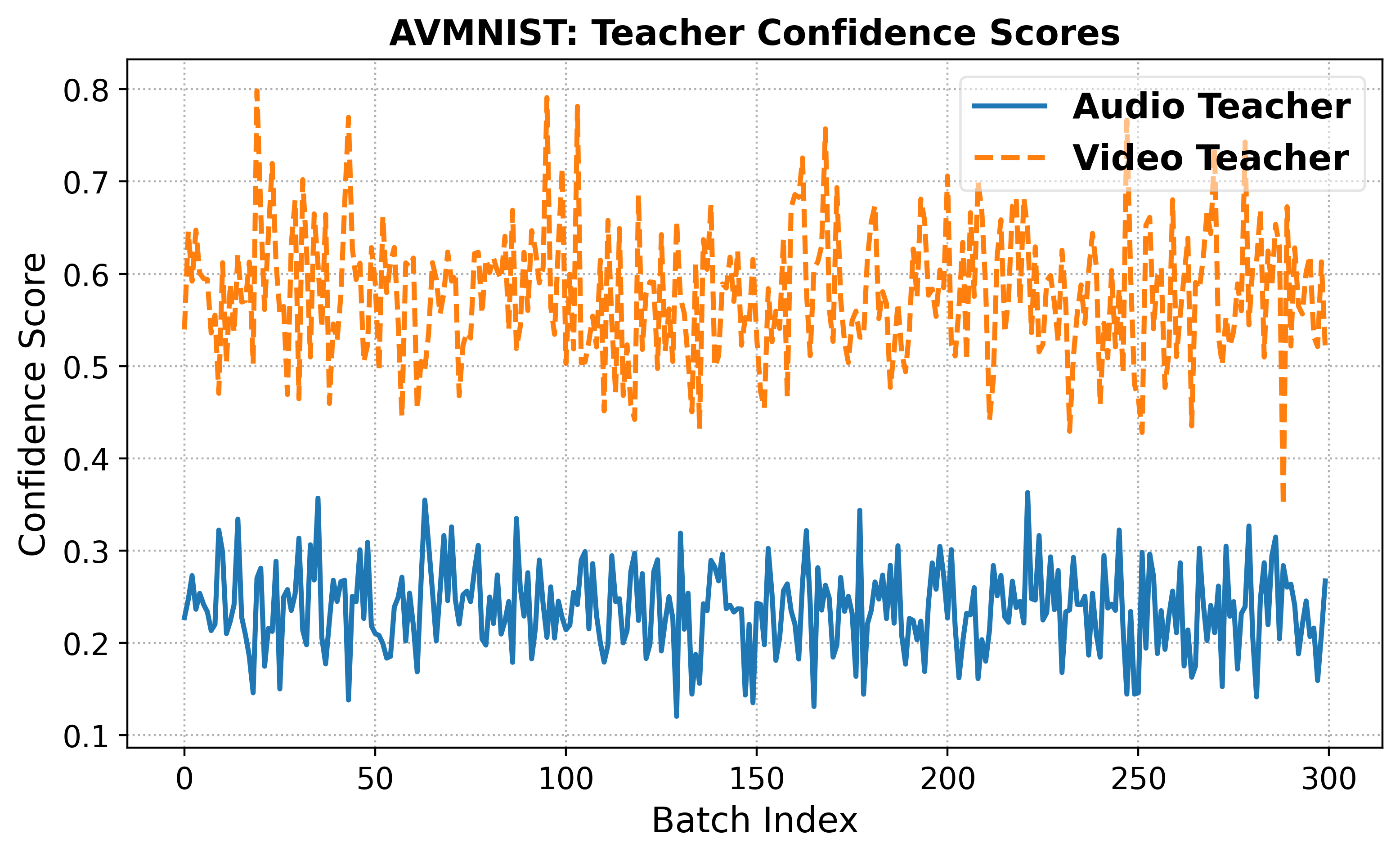}
        \caption{AV-MNIST}
        \label{fig:sub2}
    \end{subfigure}%
    \begin{subfigure}{0.25\textwidth}
        \includegraphics[width=\linewidth]{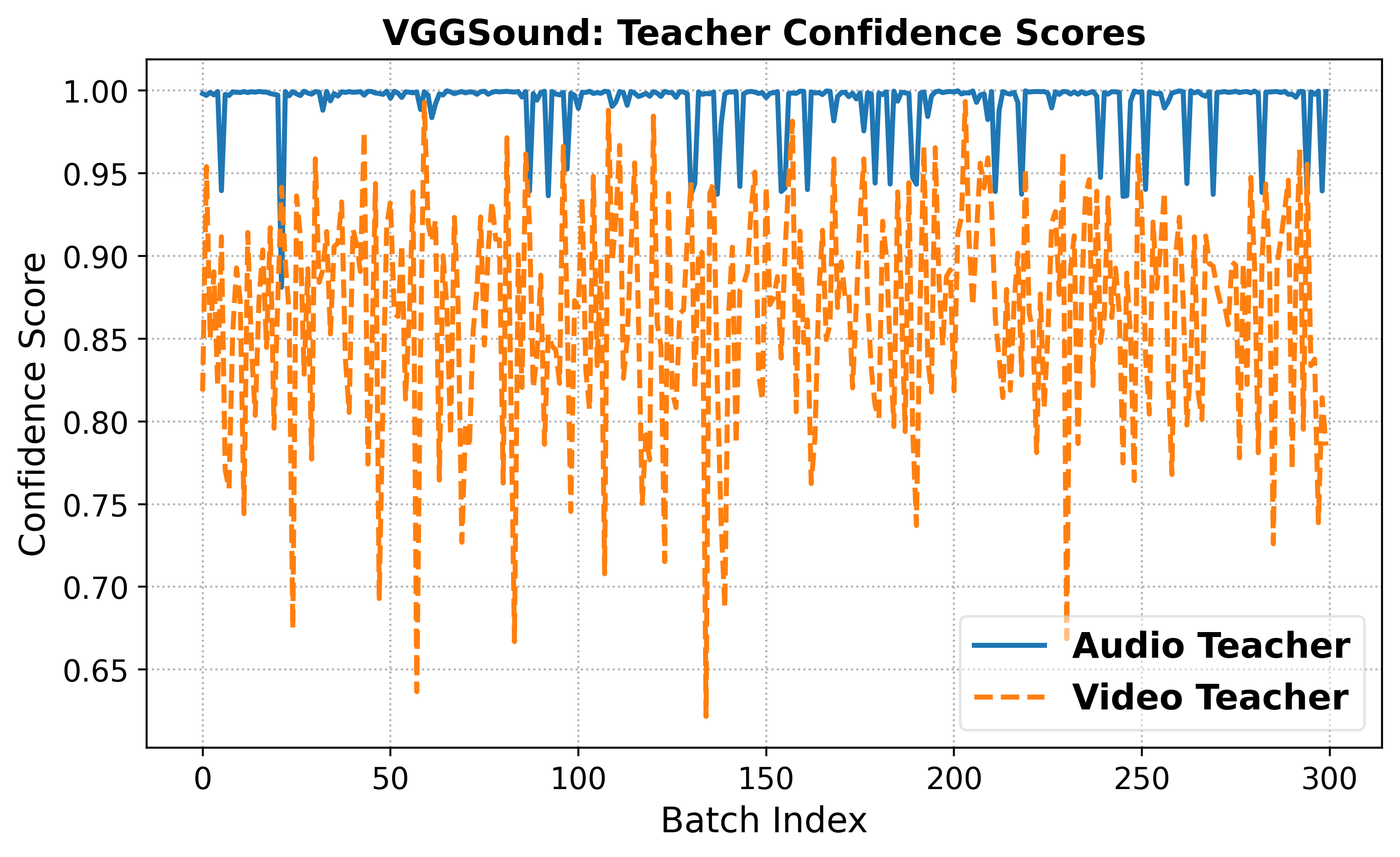}
        \caption{VGGSound}
        \label{fig:sub3}
    \end{subfigure}%
    \begin{subfigure}{0.25\textwidth}
        \includegraphics[width=\linewidth]{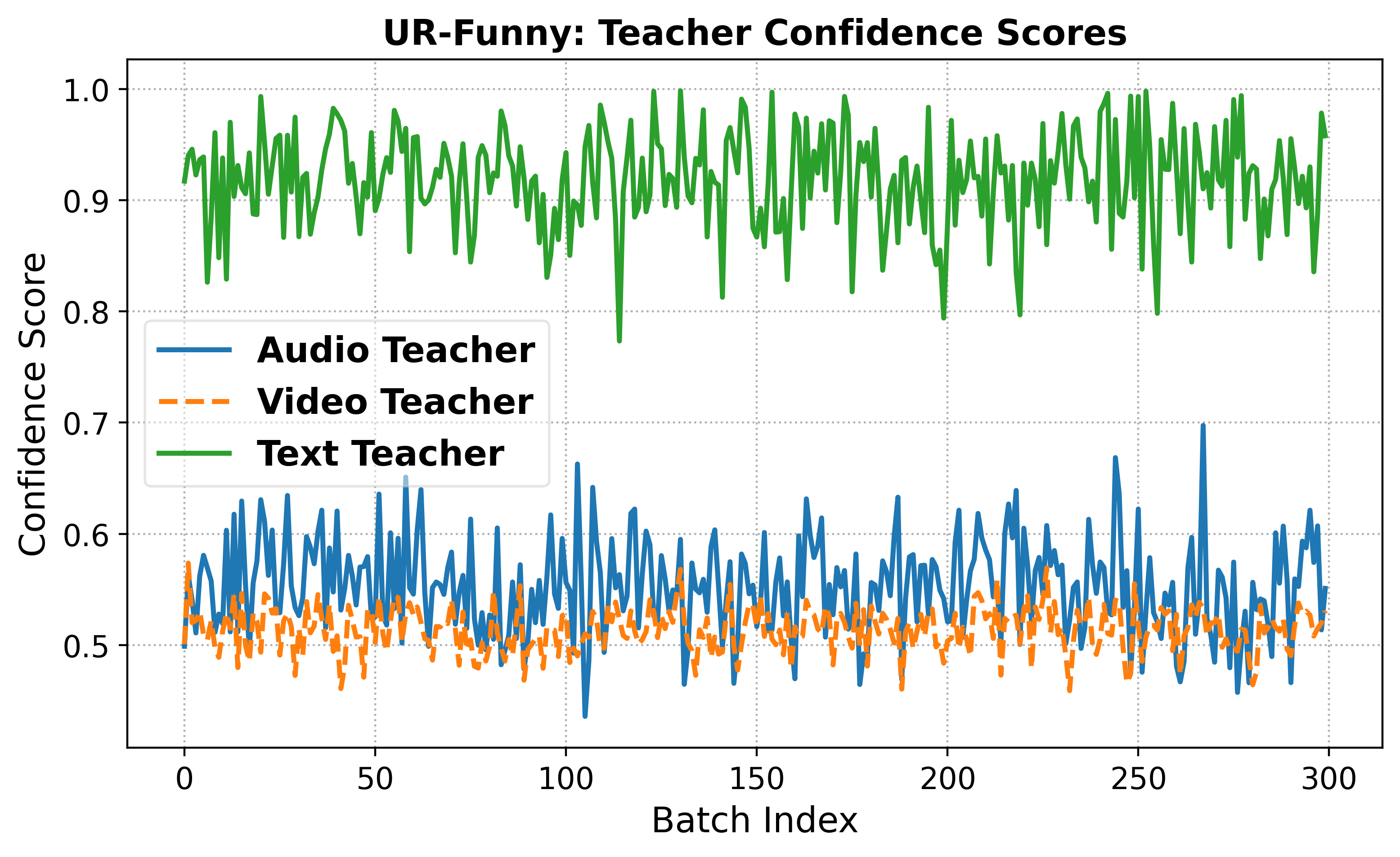}
        \caption{UR-Funny}
        \label{fig:sub4}
    \end{subfigure}
    \caption{\textbf{Unimodal teacher confidence scores across multimodal datasets.} Each line is the confidence of a specific modality (audio, visual, or text). Modality bias on all datasets, with higher scores for one modality, motivates our use of sequential modality prioritization.}
    \label{fig:scores}
\end{figure*}

% We formulate the G\textsuperscript{2}D loss function($\mathcal{L}_{\text{G\textsuperscript{2}D}}$) in such a way that we add unimodal feature loss to reduce modality competition~\cite{umt}. We further add the multimodal student logit loss objective from unimodal teachers to promote imbalance mitigation in multimodal learning. 

% 1. \textbf{Multimodal Student Loss ($\mathcal{L}_{S}$)}: We use a multimodal supervised loss function ($\mathcal{L}_S$) for the student model, as given in ~\Cref{eq:student_loss}, to ensure the model learns the correct class labels.
1. \textbf{Multimodal Student Loss ($\mathcal{L}_{S}$)} is a supervised loss to map multimodal inputs $x_i$ to the label $y_i$:

\begin{equation}
\mathcal{L}_{S} = \mathbb{E}_{(x, y) \sim [\mathcal{D}]_{m=1}^k} \left[ \ell(p, y) \right]
\label{eq:student_loss}
\end{equation}
where $\ell$ is the cross-entropy loss ($\ell(p,y) \! = \! -\sum_{w=1}^{C} y_wlog(p)$) for \emph{C}-class classification tasks with $p \! = \!  l_s(x; \theta_s)$ or mean squared error ($\ell(p,y) \! = \!  \frac{1}{N}\sum_{i=1}^{N} (p - y)^2$) for regression tasks with $p \! = \!  \sigma(l_s(x; \theta_s))$, where $\sigma$ is the sigmoid function.

% 2. \textbf{Feature Distillation Loss} ($\mathcal{L}_{\text{feat}}$): We add the feature distillation loss, as given in~\Cref{eq:feat_loss}, to encourage $m$ encoders of the student model to learn feature representations similar to those of $m$ teacher models:
2. \textbf{Feature Distillation Loss} ($\mathcal{L}_{\text{feat}}$): To prevent the student model from discarding information from weaker modalities, we impose an L2-based feature alignment loss that minimizes the discrepancy between multimodal student and unimodal teachers' feature representations:
\begin{equation}
\mathcal{L}_{\text{feat}}^{m} = \mathbb{E}_{x \sim \mathcal{D}} \left[ \| \phi_s^m(x^m; \theta_s^m) - \phi_t^m(x^m; \theta_t^m) \|^2 \right]
\label{eq:feat_loss}
\end{equation}
where $\phi_s^m$ and $\phi_t^m$ are student and teacher features, respectively, as functions of input $x^m$ and encoder parameters.

% 3. \textbf{Logits Distillation Loss ($\mathcal{L}_{\text{logit}}$)}: We propose a new loss function that aligns the student model's logits with those of the teacher models using the Kullback-Leibler divergence \cite{hinton}. We include logits in our multimodal KD as they offer greater semantic significance than deep features.
3. \textbf{Logits Distillation Loss ($\mathcal{L}_{\text{logit}}$)}: Logit-based distillation enables the student model to capture class-level relationships and decision boundaries defined by the teacher. We integrate a new logit-based distillation using Kullback-Leibler (KL) divergence \cite{hinton} that learns the distribution from unimodal teachers to the multimodal student:
\begin{equation}
\mathcal{L}_{\text{logit}}^{m} = \mathbb{E}_{x \sim \mathcal{D}} \left[ \text{KL}\left( \sigma(l_t^m(x^m; \theta_t^m)) \| \sigma(l_s(x; \theta_s)) \right) \right]
\end{equation}
where $\sigma$ denotes the softmax function. 
%Logit-based distillation enables the student model to capture class-level relationships and decision boundaries defined by the teacher, complementing feature-based alignment in knowledge transfer.

We define the G\textsuperscript{2}D loss as:
\begin{equation}
\mathcal{L}_{\text{G\textsuperscript{2}D}} = \mathcal{L}_{S} + \alpha \sum_{m=1}^{k} \mathcal{L}_{\text{feat}}^{m} + \beta \sum_{m=1}^{k} \mathcal{L}_{\text{logit}}^{m}
\label{eq:TL}
\end{equation}
where  $\alpha$ and $\beta$ are weighting coefficients for the feature loss and the logit loss, respectively. This formulation enables the multimodal student model to leverage the strengths of each unimodal teacher model effectively. Feature distillation loss ensures that the student retains modality-specific representations, while logit distillation loss aligns the student’s predictions with teacher distributions, capturing inter-class and intra-class dependencies. Although this new learning objective enables maximization of quality, multimodal models can still be biased to dominant modalities as they optimize all modalities simultaneously. We introduce an adaptive training strategy based on unimodal confidence scores to mitigate modality imbalance.

\subsection{Quantifying Modality Confidence}

%To inform the gradient modulation process, we introduce a scoring mechanism that quantifies the confidence of each unimodal teacher model. The confidence score for each modality \( m \), denoted as \( \rho_{t}^{m} \), is computed based on the softmax output of the teacher model \( T_m \).

Scoring mechanisms are widely used to measure the contributions of individual modalities in multimodal learning~\cite{ogm-ge, agm}. As unimodal models are outside the bounds of modality imbalance, we utilize their confidence in determining the imbalance ratio. Our scoring module in G\textsuperscript{2}D quantify the confidence $\rho$ of each unimodal teacher $T^m$ as the batch-wise average of their softmax function: 

%For a given sample \( x_i \), the teacher model \( T_m \) generates logits \( l_{t}^{m} \). The softmax function is applied to these logits to obtain the predicted probabilities for each class. The confidence score \( s_{t}^{m} \) is defined as the average probability assigned to the ground truth label \( y_i \) across all samples in a batch:

\begin{equation}
\rho_{t}^{m} = \frac{1}{|\mathcal{B}^m|} \sum_{(x_i^m, y_i^m) \in \mathcal{B}^m} \text{Softmax}(l_{t}^{m}(x_i^m;\theta^m))[y_i^m],
\label{eq:score}
\end{equation}
where \( |\mathcal{B}^{m}| \) represents the number of $m$ modality data samples in the batch, and \( \text{Softmax}(l_{t}^{m}(x_i^m;\theta^m))[y_i^m] \) is the probability assigned to the ground truth label \( y_i^m \) by the teacher model \( T^m \) for the sample \( x_i \).

The score \(\rho_{t}^{m}\) serves as an indicator of how confident the unimodal teacher \( T^m \) is in predicting the correct label for the given batch. 
% A higher score indicates greater confidence, and $m$ is a dominating modality, which is then used to guide the gradients in the student model, dynamically adjusting training to mitigate the dominance of any single modality.
A higher score indicates greater confidence, signifying that modality $m$ dominates in multimodal training. This information is then used to guide gradient updates in the student model, dynamically adjusting training to mitigate the dominance of any single modality.

\subsection{Modulating Gradients with Sequential Modality Prioritization (SMP)}
%\arun{Should we write this from the teacher model perspective, instead of multimodal perspective?} In particular, we note that these modality performance discrepancies are based on the modality confidence scores $\rho^m$ of each independently trained teacher model.
Multimodal datasets often undergo modality overfitting during training~\cite{g-blending} and give priority only to dominant modalities, limiting the optimization of weaker modalities. Using the modality scores $\rho^m$, we find this trend on all four classification datasets, as shown in \Cref{fig:scores}. It is evident that one modality consistently exhibits higher confidence scores, indicating a modality imbalance and dominance over other modalities in the given downstream task. We also note that this pattern is not grounded to a specific modality. For example, in CREMA-D (\Cref{fig:sub1}), the audio teacher has higher scores than the video teacher, and in UR-Funny (\Cref{fig:sub4}), the text modality remains dominant over audio and visual inputs. These findings align with previous research that identifies modality bias as a prevalent issue in multimodal datasets \cite{modality-bias1, modality-bias2, modality-bias3}. This further leads to modality imbalance in multimodal learning, as analyzed in \Cref{fig:multimodal-imbalance}. To address this issue, we hypothesize the following:

%  our analysis across multiple datasets reveals significant discrepancies in confidence scores among unimodal teacher models.

\textbf{Hypothesis 1.} \textit{Leveraging the confidence scores of unimodal models to determine less confident modalities and sequentially prioritizing them during multimodal training can mitigate modality imbalance.}\label{hypothesis}

To test this hypothesis, we propose \emph{sequential modality prioritization} strategy for the multimodal training in our proposed G\textsuperscript{2}D framework. For each training iteration $q$, we rank confidence scores of all teacher models $\rho_t^m$. If this ranked list is given as $\pi_t$, then $\pi_t[1]$ corresponds to the least confident modality, and $\pi_t[k]$ corresponds to the most dominant modality. We aim to generate an automatic training schedule process to identify the set of prioritized modalities $\mathcal{M}_q$ based on the modality confidence scores $\pi_t$, as given in \Cref{eq:mod_ranking}.
%We define a hyperparameter \(\tau_j\), where \(\tau_j\) represents the number of epochs to run the \(j\)-th prioritized modality, and we define our prioritization schedule in \Cref{eq:mod_ranking}.
%  let the confidence scores for each modality \(m \in \{m_1, \dots, m_k\}\) be calculated as \(s_{t}^{m}\) using \Cref{eq:score}. These scores are ranked in ascending order, represented by a permutation \(\pi_t\), such that \(\pi_t(1)\) corresponds to the least confident modality and \(\pi_t(k)\) corresponds to the most confident modality.
\begin{equation}
\mathcal{M}_q = 
\begin{cases}
\{\pi_t[1]\} & \text{for } 1 \leq e \leq \tau_1 \\
\{\pi_t[2]\} & \text{for } \tau_1 < e \leq \tau_1 + \tau_2 \\
\vdots \\
\{\pi_t[k-1]\} & \text{for } \sum_{j=1}^{k-2} \tau_j < e\leq \sum_{j=1}^{k-1} \tau_j \\
% \{\pi_t[1], \dots, \pi_t[k]\} & \text{for } e > \sum_{j=1}^{k-1} \tau_j\\
\{\pi_t[1], \dots, \pi_t[k]\} & \text{for } \sum_{j=1}^{k-1} \tau_j  < e \leq \sum_{j=1}^{k} \tau_j
\end{cases}
\label{eq:mod_ranking}
\end{equation}
where $e$ is the training epoch, and $\tau_j$ is the hyperparameter set to denote the number of epochs for optimizing \(j\)-th prioritized modality, where $j$ is the index of the ranked list $\pi_t$. The modulation coefficients \(\kappa_{q}^m\) for each modality \(m\) identifies the modality to optimize in multimodal learning, as given in \Cref{eq:mod_coeff}.

\begin{equation}
\kappa_{q}^{m} = 
\begin{cases}
1 & \text{if modality } m \in \mathcal{M}_q, \\
0 & \text{otherwise},
\end{cases}
\label{eq:mod_coeff}
\end{equation}

%where \(\mathcal{M}_q\) denotes the set of prioritized modalities at iteration \(q\). 

The gradient update for the parameters \(\theta_{q}^m\) of modality \(m\) for iteration $q$ in the multimodal student model \(S\) is then:
\begin{equation}
\theta_{q+1}^{m} = \theta_{q}^{m} - \eta \cdot \kappa_{q}^{m} \cdot \mathbb{E}_{(x_i, y_i) \sim \mathcal{D}} \left[ \frac{\partial \mathcal{L}_{\text{G\textsuperscript{2}D}}(x_i, y_i)}{\partial \theta_{q}^{m}} \right]
\label{eq:grad_update}
\end{equation}
where \(\eta\) is the learning rate, and \(\mathcal{L}_{\text{G\textsuperscript{2}D}}\) represents the total loss function as defined in \Cref{eq:TL}.

During each epoch range \(\tau_j\), only the corresponding modality \(\pi_t(j)\) is assigned \(\kappa_{q}^{m} = 1\), while all others are set to \(0\), ensuring that the prioritized modality receives full gradient updates. After the prioritized phases for all less confident modalities, the most confident modality (\(\pi_t(k)\)) is trained jointly with all other modalities, with \(\kappa_{q}^{m} = 1\) for all \(m\). This sequential prioritization strategy allows each modality to lead the learning process in turn, thereby mitigating persistent modality dominance. We summarize the complete multimodal training with G\textsuperscript{2}D in \Cref{alg:g2d_smp}. 
%where we treat each suppressed modality as a dominant modality to balance the optimization of all modalities.

%\arun{Should we talk about results here itself? Instead, shall we describe the Algorithm?}
%This sequential prioritization strategy allows each modality to lead the learning process in turn, thereby mitigating persistent modality dominance. The final collaborative training stage ensures a well-balanced multimodal representation without excessive bias from any single modality. Empirical results in \Cref{tab:av-results} and \Cref{tab:avt-results} demonstrate the effectiveness of this approach, as it outperforms both standard multimodal training and state-of-the-art methods, effectively mitigating modality imbalance and enhancing overall performance.

\begin{algorithm}[H]
\caption{Multimodal Learning with G\textsuperscript{2}D} 
\label{alg:g2d_smp}
\begin{algorithmic}[1]
\State \textbf{Input:} $\mathcal{D} = \{(x_i, y_i)\}_{i=1}^N$, unimodal teachers $\{T^m\}_{m=1}^{k}$, multimodal student $S$, iterations $q$
\State Initialize student model $S$ with random weights
\State Load pre-trained weights for teacher models $\{T^m\}_{m=1}^{k}$

\For{iteration $i = 0, \dots, q-1$}
    \State Sample a fresh mini-batch $(x, y)$ from $\mathcal{D}$
    \State Feed-forward the batch through $S$ to obtain $\{f_{s}^{m}\}_{m=1}^{k}$ and $l_{s}$
    \State Feed-forward each modality of a batch through $\{T^m\}_{m=1}^{k}$ to get $\{f_{t}^{m}\}_{m=1}^{k}$ and $\{l_{t}^{m}\}_{m=1}^{k}$

    \State Compute \(\mathcal{L}_{\text{G\textsuperscript{2}D}}\) loss using Eq.~(\ref{eq:TL})

    \State Calculate \(\rho_{t}^{m}\) for each modality using Eq.~(\ref{eq:score})

    \State Calculate $\kappa^m$ using Eq.~(\ref{eq:mod_coeff})

    \State Find modality gradients $\frac{\partial \mathcal{L}_{\text{G\textsuperscript{2}D}}}{\partial \theta^{m}}$ and update parameters \(\theta^m\) for each modality $m$ using Eq.~(\ref{eq:grad_update})
\EndFor

\State \Return Trained multimodal student model $S$

\end{algorithmic}
\end{algorithm}

\begin{table*}
\centering
\large
\caption{Performance comparison on various audio-visual datasets reported in accuracy (\%). "Multi" represents the evaluation of the multimodal student model, while "Audio" and "Video" rows indicate the performance of modality-specific encoders within the multimodal model. $T_a$ and $T_v$ denote unimodal evaluations for the audio and video teacher models, respectively.}
\label{tab:av-results}
\setlength{\tabcolsep}{4pt}       % Reduces column padding
\resizebox{\linewidth}{!}{%
\begin{tabular}{cc|ccc|ccccccccc|c|c} 
\toprule
\multicolumn{2}{c|}{\textbf{Dataset}} & \multicolumn{1}{c}{$T^a$} & $T^v$ & \multicolumn{1}{c|}{\begin{tabular}[c]{@{}c@{}}\textbf{Joint-}\\\textbf{Train}\end{tabular}} & \multicolumn{1}{c}{\textbf{MSES}} & \textbf{MSLR} & \textbf{AGM} & \textbf{PMR} & \begin{tabular}[c]{@{}c@{}}\textbf{OGM-}\\\textbf{GE}\end{tabular} & \textbf{MLA} & \begin{tabular}[c]{@{}c@{}}\textbf{MM}\\\textbf{Pareto}\end{tabular} & \begin{tabular}[c]{@{}c@{}}\textbf{Recon}\\\textbf{Boost}\end{tabular} & \multicolumn{1}{c|}{\textbf{DLMG}} & \textbf{UMT}  & \multicolumn{1}{c}{\begin{tabular}[c]{@{}c@{}}\textbf{G\textsuperscript{2}D}\\\textbf{(Ours)}\end{tabular}}  \\ 
\cmidrule(lr){1-16}
\multirow{3}{*}{CREMA-D}  & Audio     & 61.69                      & -     & 59.95                                                                                       & 54.86                              & 54.86         & 48.58        & 49.19        & 58.60                                                              & 59.27        & 65.46                                                                & 57.71                                                                  & 54.37                             & 61.02         & 56.45                                                                                                         \\
                          & Video     & -                          & 76.48 & 27.42                                                                                       & 22.57                              & 26.31         & 57.85        & 23.25        & 49.06                                                              & 64.91        & 55.24                                                                & 65.21                                                                  & 70.89                             & 25.40         & 72.72                                                                                                         \\
                          & Multi     & -                          & -     & 67.47                                                                                       & 60.99                              & 64.42         & 78.48        & 59.13        & 72.18                                                              & 79.70        & 75.13                                                                & 79.82                                                                  & \uline{83.62}                     & 67.61         & \textbf{85.89}                                                                                                \\ 
\cmidrule(lr){1-16}
\multirow{3}{*}{AV-MNIST} & Audio     & 42.69                      & -     & 16.05                                                                                       & 27.50                              & 22.72         & 38.90        & 37.60        & 24.53                                                              & 42.26        & 42.11                                                                & 41.50                                                                  & 41.99                             & 31.55         & 39.10                                                                                                         \\
                          & Video     & -                          & 65.44 & 55.83                                                                                       & 63.34                              & 62.92         & 63.65        & 58.50        & 55.85                                                              & 65.30        & 65.26                                                                & 64.28                                                                  & 65.04                             & 64.08         & 65.09                                                                                                         \\
                          & Multi     & -                          & -     & 69.77                                                                                       & 70.68                              & 70.62         & 72.14        & 71.82        & 71.08                                                              & 65.32        & \uline{72.63}                                                        & 72.47                                                                  & 72.14                             & 72.33         & \textbf{73.03}                                                                                                \\ 
\cmidrule(lr){1-16}
\multirow{3}{*}{VGGSound} & Audio     & 43.39                      & -     & 39.22                                                                                       & 39.57                              & 39.10         & 38.15        & 26.30        & 37.96                                                              & 37.56        & 42.44                                                                & 42.35                                                                  & 41.54                             & 42.12         & 39.43                                                                                                         \\
                          & Video     & -                          & 32.32 & 18.70                                                                                       & 17.85                              & 18.66         & 25.65        & 7.12         & 22.64                                                              & 32.02        & 17.94                                                                & 18.12                                                                  & 23.65                             & 23.77         & 29.88                                                                                                         \\
                          & Multi     & -                          & -     & 50.97                                                                                       & 50.76                              & 50.98         & 47.11        & 33.07        & 51.45                                                              & 51.65        & 49.69                                                                & 50.97                                                                  & 52.74                             & \uline{53.78} & \textbf{53.82}                                                                                                \\
\bottomrule
\end{tabular}
}
\end{table*}

\section{Experiments} 

We evaluate the G\textsuperscript{2}D framework on the basis of the following questions: \textbf{Q1:} How does G\textsuperscript{2}D compare to state-of-the-art methods for addressing modality imbalance and overall multimodal performance in supervised tasks? \textbf{Q2:} How does G\textsuperscript{2}D influence the modality gap in multimodal learning?, \textbf{Q3:} Does G\textsuperscript{2}D enhance feature space alignment between unimodal and multimodal models?, \textbf{Q4:} How SMP is influencing the multimodal learning?, and \textbf{Q5:} Which fusion and suppression techniques are the best for G\textsuperscript{2}D?

\subsection{G\textsuperscript{2}D Evaluation}
\subsubsection{Experimental Setup}
\textbf{Datasets.}
We chose five multimodal classification datasets and one regression dataset. \textbf{CREMA-D} \cite{cremad} is an audio-visual dataset for speech emotion recognition with six emotion classes. \textbf{AV-MNIST} \cite{avmnist} is a synthetic dataset with PCA-projected MNIST images and audio spectrograms for ten-digit classes. \textbf{VGGSound} \cite{vggsound} is a large-scale audio-visual dataset with 309 classes of everyday audio events, featuring video clips of 10 seconds each. \textbf{UR-Funny} \cite{urfunny} is a binary classification dataset for humor detection that incorporates text, visual gestures, and acoustic modalities. \textbf{IEMOCAP} \cite{iemocap} is an audio-visual-text dataset for emotion recognition in dyadic conversations. \textbf{MIS-ME} \cite{misme} is a regression dataset containing raw soil patch images and corresponding meteorological data for soil moisture estimation. To the best of our knowledge, we are the \emph{first} to evaluate the modality imbalance in a multimodal regression setting and use tabular data for multimodal learning problems.
%, using facial and vocal emotional expressions

\textbf{Baselines.}
We compare G\textsuperscript{2}D with ten state-of-the-art methods that address modality imbalance, including MSES \cite{mses}, MSLR \cite{mslr}, AGM \cite{agm}, PMR \cite{pmr}, OGM-GE \cite{ogm-ge}, MLA \cite{mla}, MM-Pareto \cite{mm-pareto}, ReconBoost\cite{reconboost}, DLMG\cite{dlmg}, and UMT \cite{umt}. We compare the baseline methods across four datasets that include audio, visual, and text modalities. For the regression task, we compare our approach to MISME \cite{misme}, which estimates soil moisture from soil patch images and meteorological data.  We evaluate the performance of each baseline in both unimodal and multimodal contexts for each modality. Please refer to \cref{baseline-supp} for more detailed baseline descriptions.
%Specifically, we report results using all available modalities for multimodal evaluation while also assessing performance of \arun{fine-tuned} multimodal model using only a single modality. 

\textbf{Backbone and Hyperparameter Settings.}  
For audio-visual datasets (CREMA-D, VGGSound, and AV-MNIST), we use ResNet-18~\cite{resnet18} as the encoder for both audio and video modalities in the teacher and student models. For the UR-Funny and IEMOCAP dataset, which involves audio, visual, and text modalities, we use a Transformer-based encoder~\cite{transformer} for each modality. For MIS-ME, we adopt MobileNetV2~\cite{mobilenetv2} as the image feature extractor and use the fully connected neural network proposed in \cite{misme} for processing tabular meteorological data. To ensure a fair comparison, we use identical backbone architectures across all baseline models and employ late fusion for training. All models are optimized using SGD with a batch size of 16 and trained on a single NVIDIA A10 GPU. 
More details on experimental settings are provided in \cref{experimental-setup-supp}.

\subsubsection{Results}
In this section, we present the accuracy($\%$) of all models with best results in \textbf{bold} and second best results \underline{underlined}.

\textbf{G\textsuperscript{2}D on Two Modalities and Audio-Visual Domain.} Table~\ref{tab:av-results} presents the following key observations:

1. Unimodal performances ($T_a$ and $T_v$) and joint-training reveal that modality imbalance is dataset-dependent. On CREMA-D and VGGSound, video performs well in the unimodal setup but becomes suppressed in multimodal training, while it is the opposite in AV-MNIST as audio is underutilized in multimodal setup. This confirms that modality imbalance is a prevalent issue in multimodal learning, leading to suboptimal fusion in joint-training. 

2. DLMG, ReconBoost, and gradient modulation methods (AGM, OGM-GE, MLA, and MMPareto) attempt to reduce modality imbalance and improve fusion. While effective, they do not fully bridge the discrepancy among modalities, as imbalance persists across datasets. G\textsuperscript{2}D surpasses all baselines, demonstrating that SMP ensures balanced optimization and better multimodal integration.  

3. To the best of our knowledge, UMT is the only knowledge distillation-based baseline addressing modality imbalance. The results show that the proposed $G^2D$ loss that distills knowledge from unimodal teachers with the dynamic training strategy using SMP gives better optimization for weak modalities to outperform UMT across all datasets.

\textbf{G\textsuperscript{2}D on Three Modalities and Text Domain.} Unlike prior approaches \cite{greedy, pmr}, G\textsuperscript{2}D is not constrained by the number of modalities. We now analyze results with a three-modality dataset on UR-FUNNY, given in Table~\ref{tab:avt-results}. In this special case of experiment, we compare our results with baseline models using all combinations of modalities. We find with joint-training that text modality is dominant in all multimodal settings and incorporating all modalities leads to the best multimodal performance. (i) \textbf{G\textsuperscript{2}D performance:} We observe that G\textsuperscript{2}D consistently outperforms methods incorporating adaptive training strategies, such as OGM-GE, MMPareto, and ReconBoost, as well as the KD-based approach UMT, demonstrating its effectiveness in mitigating modality imbalance. (ii) \textbf{Modality depression:} Another interesting finding with more than two modalities (\textbf{A-V-TXT}) is the dominant modality (text) depression across most of our baseline models. We suspect that these models give over-prioritization to weak modalities while not allowing the required optimization for dominant modalities. G\textsuperscript{2}D, on the other hand, treats all modalities fairly with the proposed SMP to reduce modality depression. 
%The results further indicate that incorporating all three modalities leads to the best multimodal performance across all methods, reinforcing the benefit of leveraging richer cross-modal interactions. 

% However, the audio modality dominates the visual modality when we do not consider the text modality.

\begin{table}[h]
\centering
\large
\caption{Accuracy (\%) on the UR-Funny dataset across all modality combinations. 
% Performance comparison on the UR-Funny dataset across audio-video (A-V), audio-text (A-TXT), visual-text (V-TXT), and audio-visual-text (A-V-TXT) combinations, reported in accuracy (\%).
% "Multi" represents the evaluation of the multimodal student model, while "Audio," "Visual," and "Text" indicate the performance of modality-specific encoders within the multimodal model. 
Unimodal teacher performance for audio, visual, and text are $61.57\%$, $58.25\%$, and $61.77\%$, respectively.}
\label{tab:avt-results}
\setlength{\tabcolsep}{4pt}       % Reduces column padding
\resizebox{\columnwidth}{!}{%
\begin{tabular}{cc|c|ccc|c|c} 
\toprule
\multicolumn{2}{c|}{\textbf{Type}} & \begin{tabular}[c]{@{}c@{}}\textbf{Joint-}\\\textbf{Train}\end{tabular} & \multicolumn{1}{c}{\begin{tabular}[c]{@{}c@{}}\textbf{OGM-}\\\textbf{GE}\end{tabular}} & \begin{tabular}[c]{@{}c@{}}\textbf{MM}\\\textbf{Pareto}\end{tabular} & \multicolumn{1}{c|}{\begin{tabular}[c]{@{}c@{}}\textbf{Recon}\\\textbf{Boost}\end{tabular}} & \textbf{UMT}  & \multicolumn{1}{c}{\begin{tabular}[c]{@{}c@{}}\textbf{G\textsuperscript{2}D }\\\textbf{(Ours)}\end{tabular}}  \\ 
\cmidrule(r){1-8}
\multirow{3}{*}{A-V}     & Audio   & 57.34                & 59.76           & 61.77             & 60.53               & 54.63         & 59.05                                  \\
                         & Visual  & 53.92                & 53.82           & 55.73             & 57.87               & 56.44         & 58.05                                  \\
                         & Multi   & 61.57                & 61.87   & 61.27             & \uline{62.07}               & 60.46         & \textbf{62.98}                         \\ 
\cmidrule(r){1-8}
\multirow{3}{*}{A-TXT}   & Audio   & 50.30                & 54.12           & 58.15             & 50.18               & 55.63         & 59.86                                  \\
                         & Text    & 57.44                & 58.35           & 58.45             & 56.98               & 57.75         & 58.85                                  \\
                         & Multi   & 62.17                & 62.47           & \uline{62.88}     & 61.06               & 62.47         & \textbf{63.28}                         \\ 
\cmidrule(lr){1-8}
\multirow{3}{*}{V-TXT}   & Visual  & 49.30                & 55.33           & 56.04             & 55.41               & 56.34         & 56.34                                  \\
                         & Text    & 51.21                & 58.95           & 59.15             & 50.94               & 53.82         & 56.04                                  \\
                         & Multi   & 62.07                & 62.98           & 61.27             & 60.07               & \uline{63.18} & \textbf{63.48}                         \\ 
\cmidrule(r){1-8}
\multirow{4}{*}{A-V-TXT} & Audio   & 55.03                & 50.30           & 58.05             & 51.65               & 50.70         & 59.15                                  \\
                         & Visual  & 54.93                & 55.73           & 56.14             & 55.26               & 54.93         & 55.94                                  \\
                         & Text    & 58.25                & 55.71           & 58.55             & 56.25               & 52.72         & 58.15                                  \\
                         & Multi   & 62.58                & \uline{63.68}   & 62.88             & 61.37               & 63.38         & \textbf{65.49}                         \\
\bottomrule
\end{tabular}
}
\end{table}

%Notably, G\textsuperscript{2}D achieves the highest accuracy in every multimodal evaluation, demonstrating its robustness in extracting and combining complementary information from diverse modalities.

\begin{figure*}[t]
    \centering
    \begin{subfigure}{0.24\textwidth}
        \includegraphics[width=0.85\linewidth]{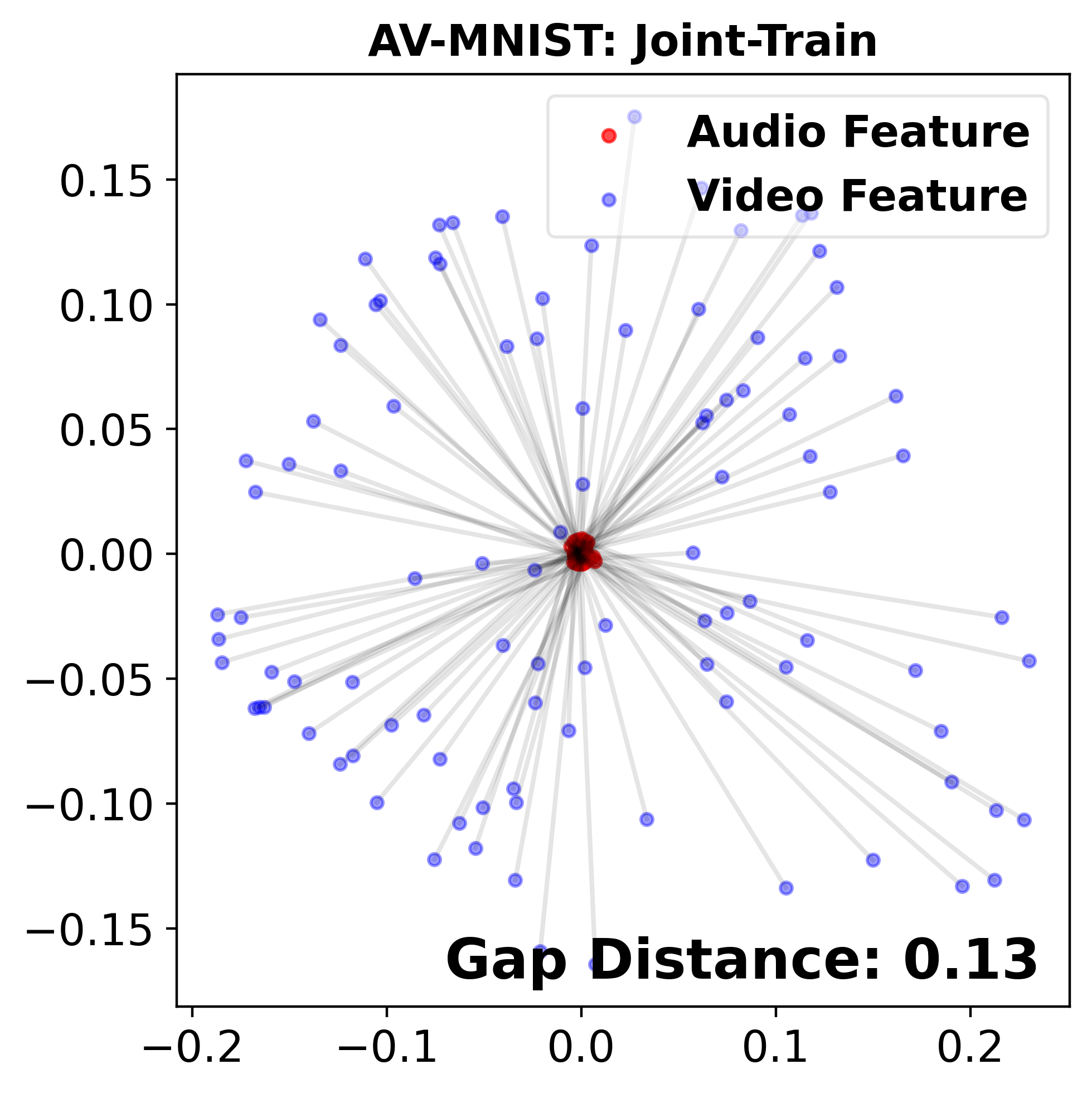}
        \caption{AV-MNIST: Joint-Train}
        \label{fig:gapsub1}
    \end{subfigure}%
    \begin{subfigure}{0.24\textwidth}
        \includegraphics[width=0.85\linewidth]{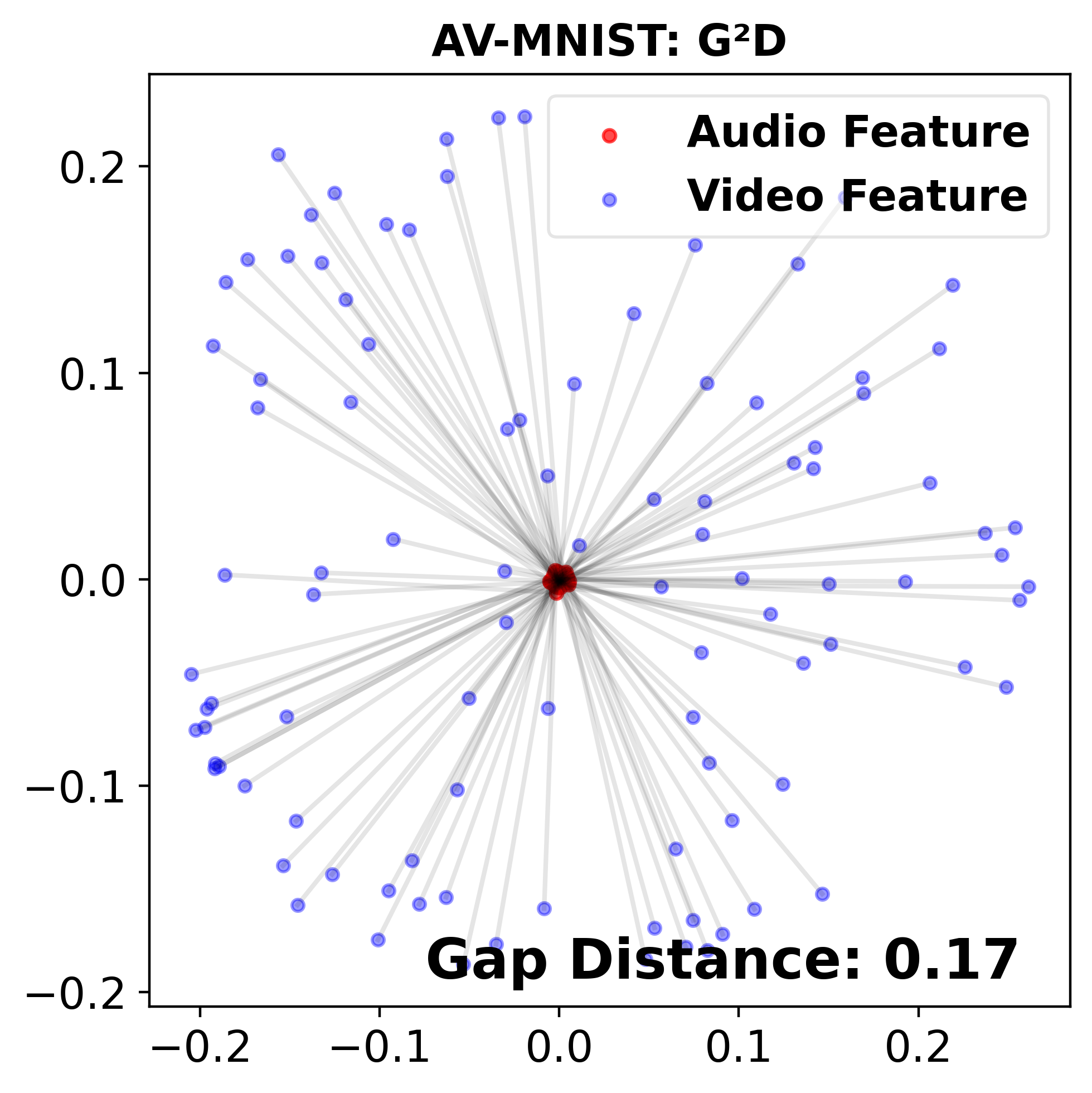}
        \caption{AV-MNIST:  G\textsuperscript{2}D}
        \label{fig:gapsub2}
    \end{subfigure}%
    \begin{subfigure}{0.24\textwidth}
        \includegraphics[width=0.85\linewidth]{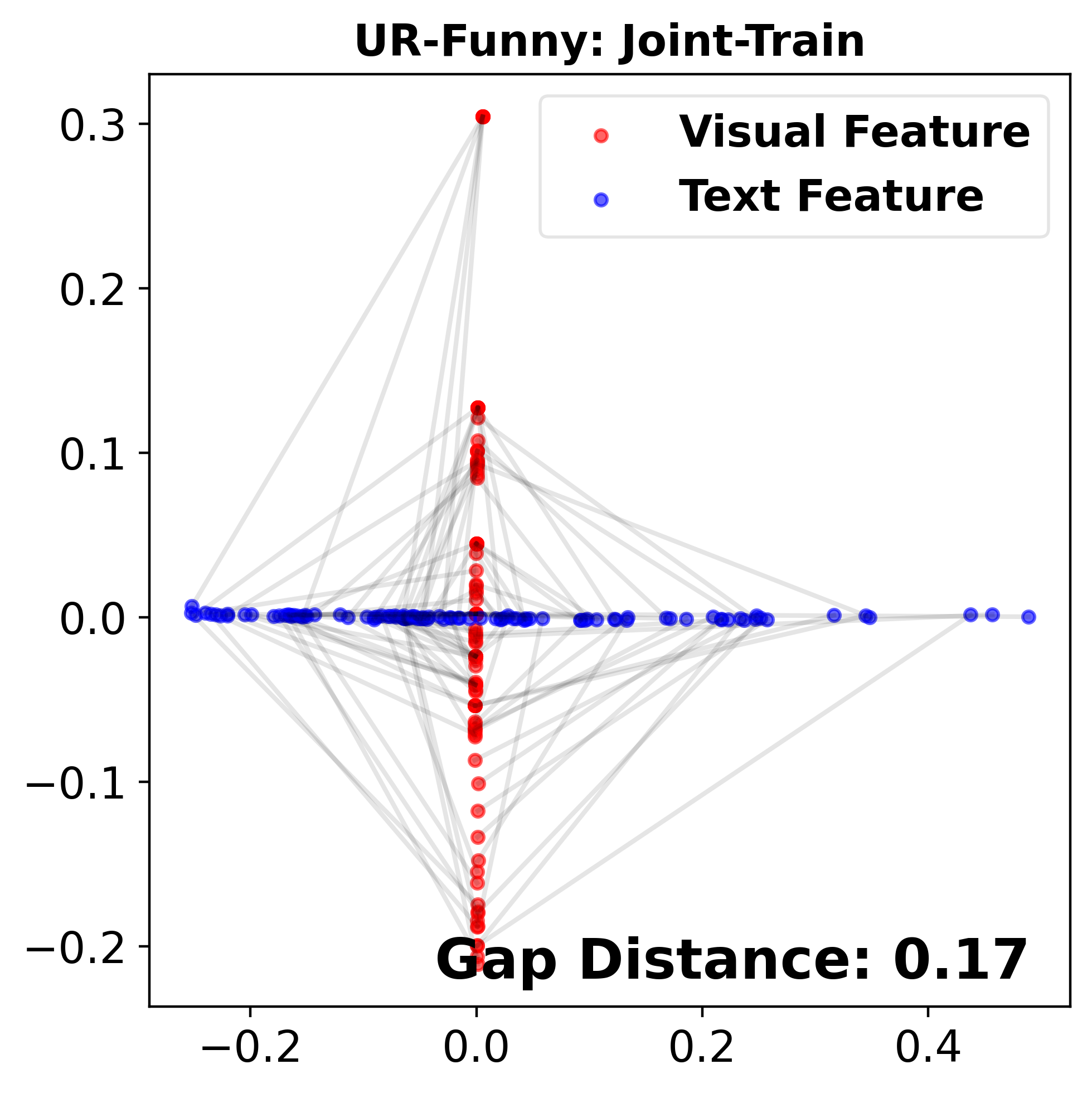}
        \caption{UR-Funny: Joint-Train}
        \label{fig:gapsub3}
    \end{subfigure}%
    \begin{subfigure}{0.24\textwidth}
        \includegraphics[width=0.85\linewidth]{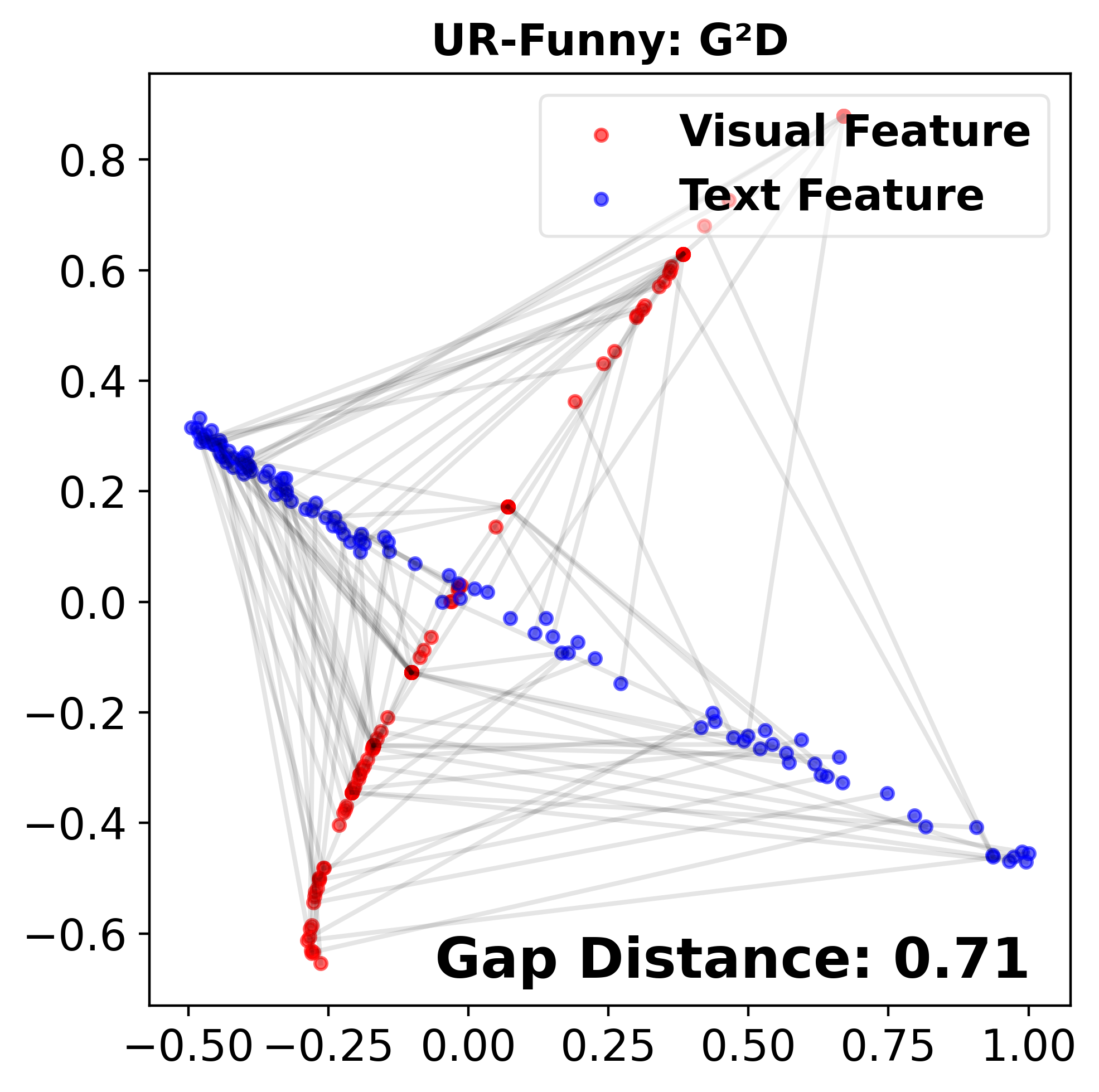}
        \caption{UR-Funny:  G\textsuperscript{2}D}
        \label{fig:gapsub4}
    \end{subfigure}
    \caption{Modality gap for AV-MNIST and UR-Funny datasets, with G\textsuperscript{2}D increasing the modality separation compared to joint-training. }
    \label{fig:modalitygap}
\end{figure*}

\textbf{G\textsuperscript{2}D on the Multimodal Regression Task.} To evaluate the robustness and real-world applicability of G\textsuperscript{2}D, we applied it to a soil moisture estimation task using an in-wild raw soil patch dataset \cite{misme}, captured via cameras. As given in \Cref{tab:regression}, we find that modality imbalance occurs in regression tasks as well. We find that G\textsuperscript{2}D outperforms the baseline method MIS-ME, indicating its effectiveness in multimodal regression tasks. 

%by leveraging improved gradient modulation to better align multimodal features.

\begin{table}[h]
\centering
\caption{G\textsuperscript{2}D for Soil Moisture Regression Task on MIS-ME Dataset with \emph{tabular} and \emph{image} Modalities}
\label{tab:regression}
\resizebox{\columnwidth}{!}{%
\begin{tabular}{c|ccc|cc} 
\toprule
\textbf{Metrics}     & \textbf{Tabular} & \textbf{Image} & \textbf{Joint-Train} & \textbf{MIS-ME} & \textbf{G\textsuperscript{2}D}  \\ 
\cmidrule(r){1-6}
MAPE                 & 15.49            & 8.22           & 14.62                & \uline{7.52}            & \textbf{7.01}                   \\ 
\cmidrule(lr){1-6}
R\textsuperscript{2} & 0.34             & 0.76           & 0.42                 & \uline{0.80}            & \textbf{0.82}                   \\
\bottomrule
\end{tabular}
}
\end{table}

\subsection{Analysis of G\textsuperscript{2}D}

% As demonstrated in prior work on modality gap \cite{modalitygap, modalitygap2}, multimodal learning leads to different modalities being embedded in distinct regions of the representation space. This modality gap can be correlated with improved model performance, as separating modalities can enhance their individual contributions during learning \cite{modalitygap2}. To gain insights into the performance improvements achieved by G\textsuperscript{2}D, we visualize the modality gap between two modalities in AV-MNIST (audio-visual) and UR-Funny (visual-text) datasets, as shown in \Cref{fig:modalitygap}. Compared to the joint-training approach (\Cref{fig:gapsub1,fig:gapsub3}), G\textsuperscript{2}D (\Cref{fig:gapsub2,fig:gapsub4}) results in a larger modality gap, signifying that different modalities become more distinguishable in the embedding space. We observe that multimodal models with text modality can achieve greater separation in the feature space. This enhanced separation between modalities facilitates more effective feature utilization, contributing to the observed improvements in downstream tasks. The visualization further underscores the effectiveness of our approach in mitigating modality bias and leveraging the complementary strengths of each modality for superior multimodal learning.

\textbf{Analyzing Modality Gap.} Prior work has shown that multimodal learning leads to distinct modality-specific embeddings, with a larger \textit{modality gap} often correlating with improved performance \cite{modalitygap, modalitygap2}. Following these insights, we visualize the modality gap in AV-MNIST (audio-visual) and UR-Funny (visual-text) datasets, as shown in \Cref{fig:modalitygap}. Compared to joint-training (\Cref{fig:gapsub1,fig:gapsub3}), G\textsuperscript{2}D (\Cref{fig:gapsub2,fig:gapsub4}) results in a more pronounced modality gap, making modalities more distinguishable in the embedding space. This separation is particularly evident in text-inclusive models, facilitating better feature utilization. These results highlight G\textsuperscript{2}D’s ability to mitigate modality bias and enhance multimodal learning by preserving modality-specific characteristics.

\begin{figure}[h]
    \centering
    \begin{subfigure}{0.50\columnwidth}
        \centering
        \includegraphics[width=0.85\textwidth]{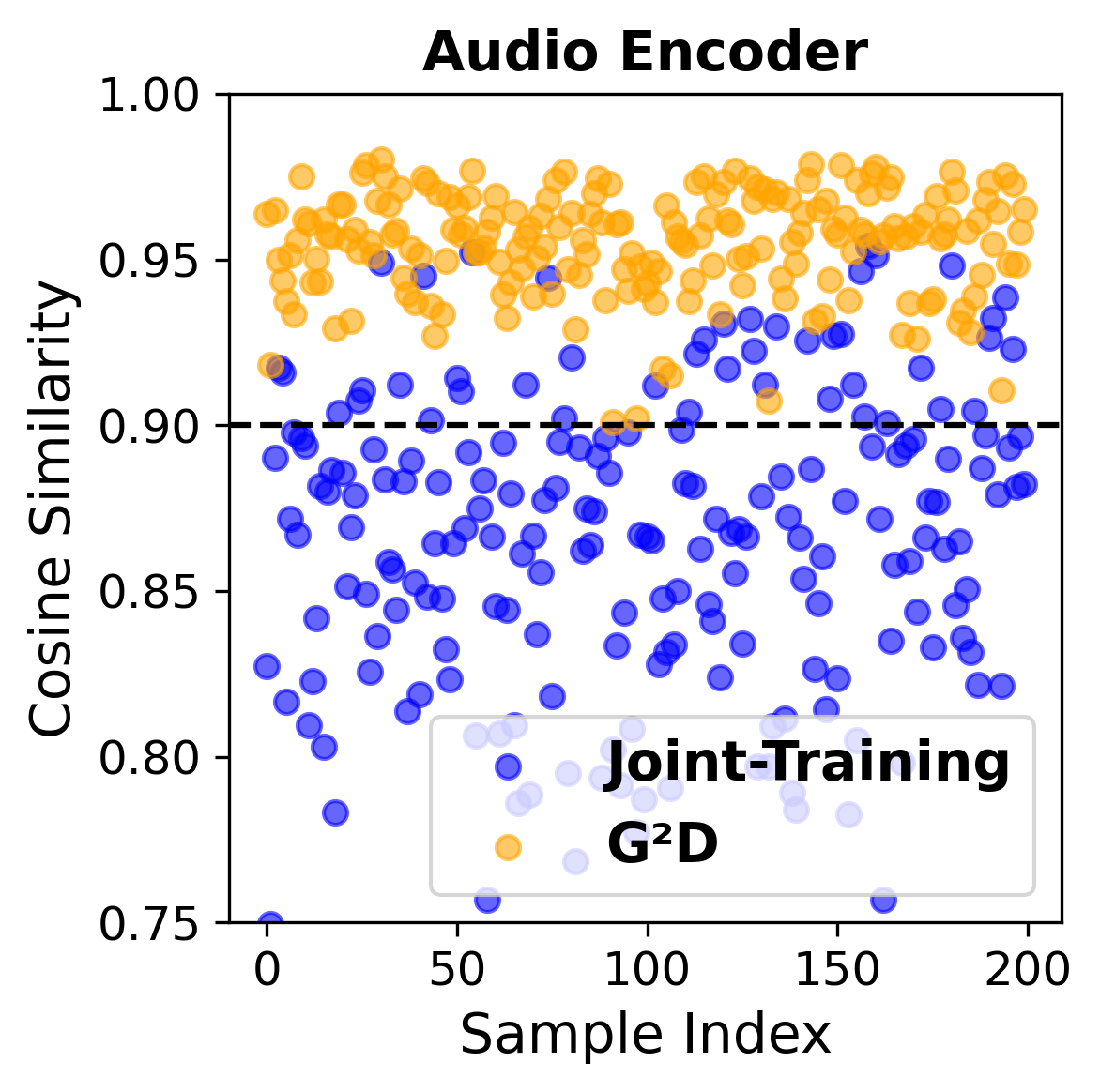}
        \caption{Audio Encoder}
        \label{fig:audio-encoder}
    \end{subfigure}
    \hspace{-0.5cm}  % Reduce the horizontal space
    \begin{subfigure}{0.50\columnwidth}
        \centering
        \includegraphics[width=0.85\textwidth]{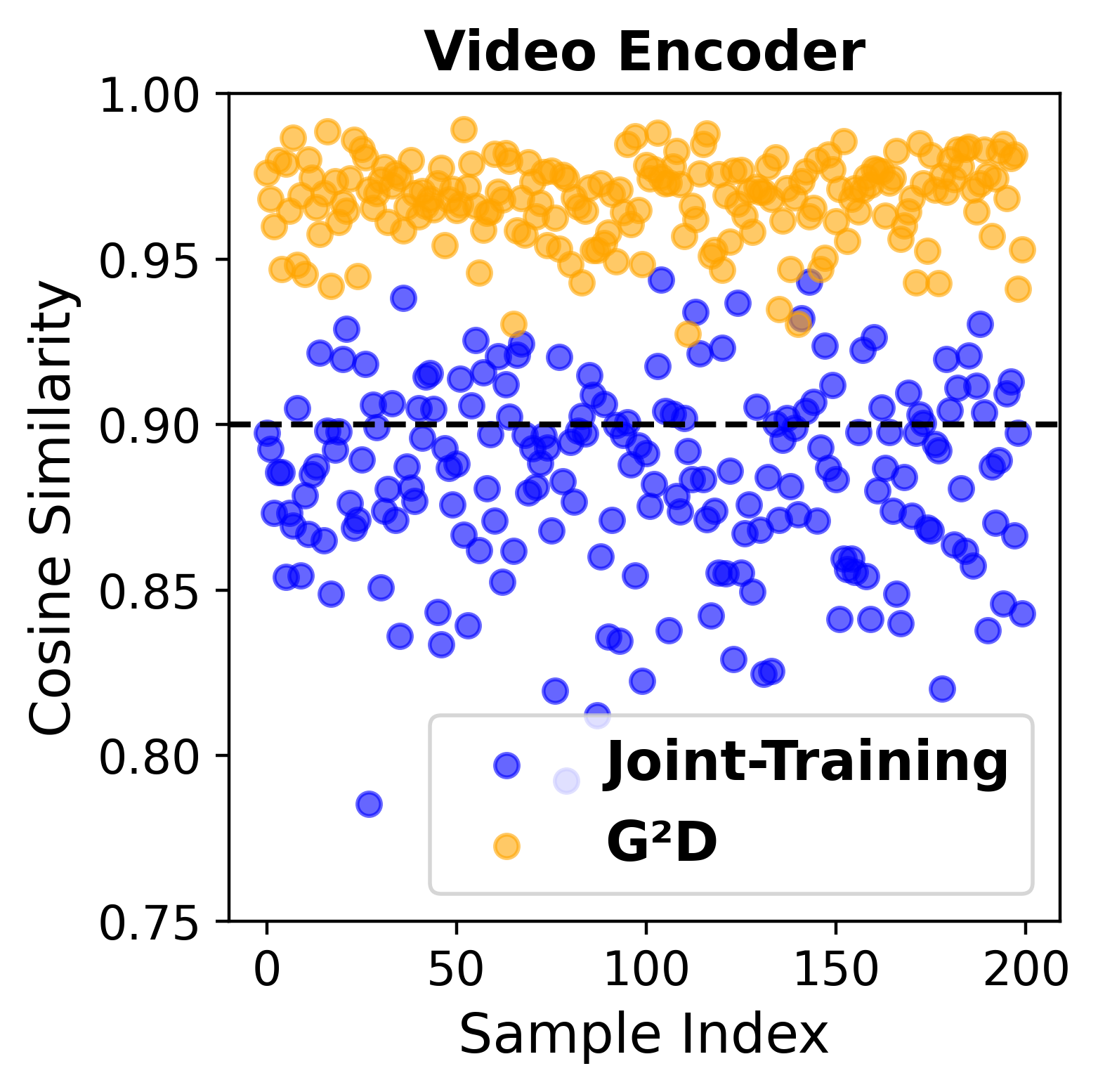}
        \caption{Video Encoder}
        \label{fig:video-encoder}
    \end{subfigure}
    \caption{Alignment between unimodal and multimodal features in the audio encoder (\Cref{fig:audio-encoder}) and the video encoder (\Cref{fig:video-encoder}).}
    \label{fig:encoder-comparison}
\end{figure}

\textbf{Analyzing Feature Alignment of G\textsuperscript{2}D.}
We first analyze the robustness of multimodal features by aligning them with unimodal teacher features for both audio and video encoders in CREMA-D. In this experiment, we use cosine similarity to measure the feature alignment. With \Cref{fig:audio-encoder} and \Cref{fig:video-encoder}, we show that the alignment of both modalities between unimodal and multimodal features is consistently higher for the G\textsuperscript{2}D compared to a simple joint-training without KD. We consider that better feature alignment in G\textsuperscript{2}D is one of the important factors in improving modality imbalance in multimodal learning.

\begin{figure}[h]
  \centering
  \includegraphics[width=0.80\linewidth]{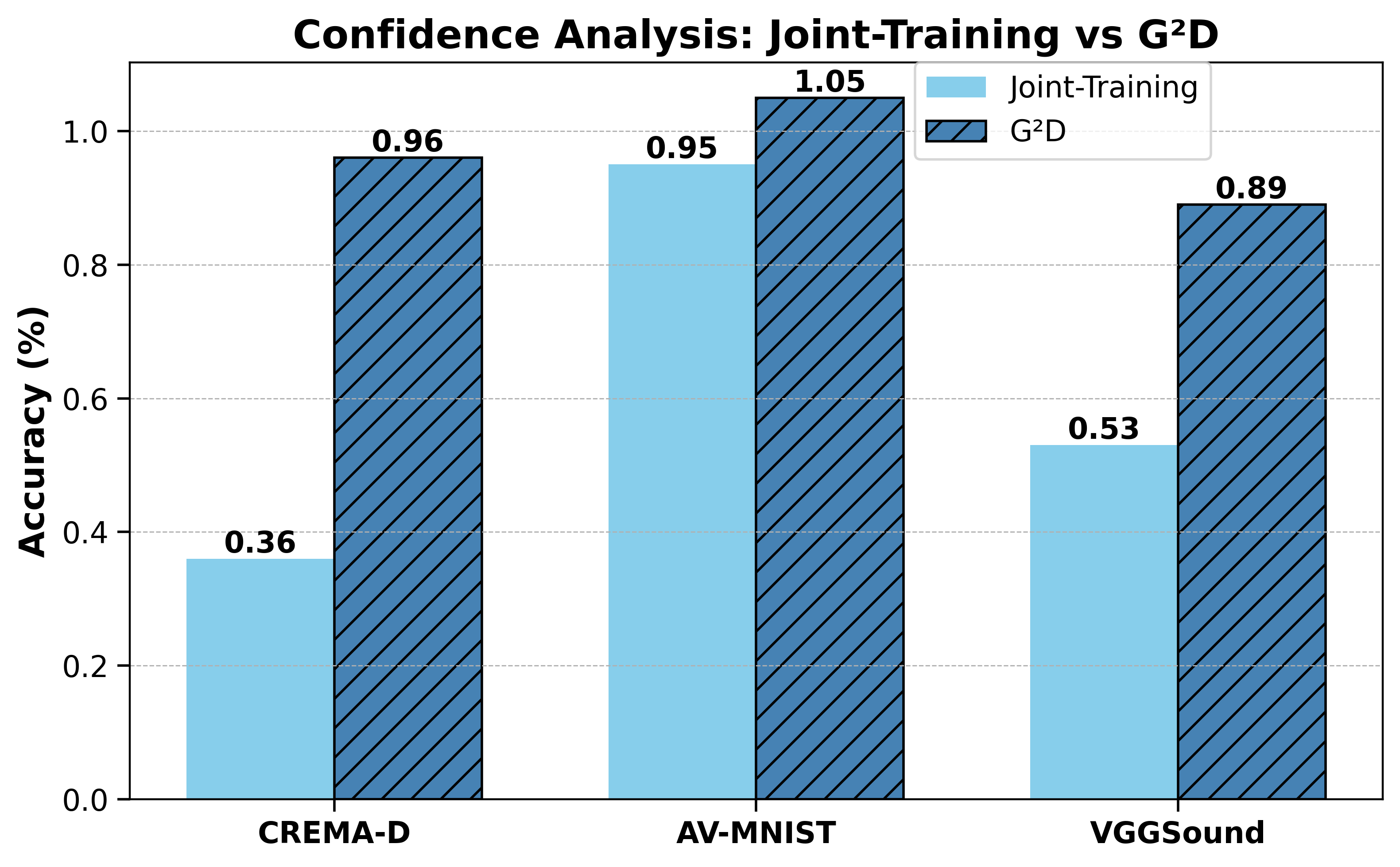}
  \caption{Analyzing Confidence Ratio of Joint-Training with G\textsuperscript{2}D}
  \label{fig:score-analysis}
\end{figure}

\textbf{Analyzing Modality Imbalance with Confidence Ratio.} To quantify the impact of modality imbalance, we compute \textit{confidence ratio} to measure the weaker modality's confidence relative to its unimodal teacher. Specifically, we first compute the average confidence score $\rho$ for the weaker modality across the entire dataset in both joint-training and G\textsuperscript{2}D. Then, we normalize these values by dividing them by the corresponding unimodal teacher's confidence score. A lower confidence ratio indicates that the weaker modality is overshadowed during training, while a higher ratio suggests that the weaker modality in multimodal setting is reaching its performance close to that of its unimodal setup. As shown in \cref{fig:score-analysis}, G\textsuperscript{2}D consistently yields higher confidence ratios than joint-training, demonstrating its ability to mitigate modality discrepancy by ensuring a more balanced optimization process and preventing weaker modalities from being suppressed in the multimodal training.

Further analysis of G\textsuperscript{2}D, including distinction from key baselines and computational cost, is in \cref{g2d-analysis-supp}.

\subsection{Ablation Study}

\begin{table}[th]
\centering
\large
\caption{Effect of SMP on Different Multimodal Methods}
\label{tab:component_results}
\resizebox{0.9\columnwidth}{!}{
\begin{tabular}{c|c|ccc} 
\toprule
\textbf{Method}                             & \textbf{SMP} & \textbf{CREMA-D} & \textbf{AV-MNIST} & \textbf{UR-Funny}  \\ 
\cmidrule(lr){1-5}
\multirow{2}{*}{Joint-Train}                & \xmark       & 67.47            & 69.77             & 62.58              \\
                                            & \checkmark   & 80.78                 & 72.51                  & 63.58                   \\ 
\cmidrule(lr){1-5}
\multirow{2}{*}{UMT}                        & \xmark       & 67.61            & 72.33             & 63.38              \\
                                            & \checkmark   & \uline{82.39}            & 72.68                   & \uline{64.59}              \\ 
\cmidrule(lr){1-5}
\multirow{2}{*}{G\textsuperscript{2}D loss} & \xmark       & 78.63            & \uline{72.76}             & 63.78              \\
                                            & \checkmark   & \textbf{85.89}   & \textbf{73.03}    & \textbf{65.49}     \\
\bottomrule
\end{tabular}
}
\end{table}

\textbf{Impact of SMP.} The results in Table~\ref{tab:component_results} demonstrate \emph{two-pronged} observations on SMP with three datasets. (i) SMP integration enhances the performance of both vanilla joint-training and the multimodal KD-based approach UMT. (ii) Incorporating SMP to G\textsuperscript{2}D loss achieves the best overall performance, outperforming all baselines, highlighting its role in mitigating modality imbalance and improving multimodal learning. This confirms the effectiveness and adaptability of SMP, not just in G\textsuperscript{2}D but also in other models.

\begin{table}[h]
\centering
\large
\caption{Performance comparison of G\textsuperscript{2}D using different fusion strategies across multiple datasets in terms of accuracy (\%). }
% "Multi" represents the evaluation of the multimodal student model, while "Audio," "Visual," and "Text" indicate the performance of modality-specific encoders. 
% The best and second-best performances for each dataset are highlighted in \textbf{bold} and \underline{underlined}, respectively.
\label{tab:fusion-results}
\resizebox{\linewidth}{!}{%
\begin{tabular}{c|cccc} 
\toprule
\textbf{Fusion} & \textbf{CREMA-D}        & \textbf{AV-MNIST}       & \textbf{VGGSound}       & \textbf{UR-Funny}  \\ 
\cmidrule(lr){1-5}
Sum             & 81.59                   & 72.70                   & 50.67                   & 63.08              \\
Concat          & 83.60                   & \uline{72.98}           & 53.40                   & 64.49              \\
FiLM \cite{film}           & 84.27                   & 72.73                   & 48.11                   & 63.48              \\
BiGated \cite{gated}        & 81.32                   & 72.89                   & 46.66                   & 63.38              \\
Cross-Attention \cite{cross-attention} & \uline{85.35}           & 72.96                   & \uline{53.58}           & \uline{65.09}     \\
Late Fusion \cite{late-fusion}     & \textbf{\textbf{85.89}} & \textbf{\textbf{73.03}} & \textbf{\textbf{53.82}} & \textbf{65.49}      \\
\bottomrule
\end{tabular}
}
\end{table}

% In \Cref{tab:fusion-results}, we evaluate G\textsuperscript{2}D using different fusion techniques. 
% \textbf{G\textsuperscript{2}D with Various Fusion Modules.} Results in \Cref{tab:fusion-results} show that Late Fusion and Concat consistently outperform other fusion techniques across most datasets, demonstrating their superior capability to thoroughly capture cross-modal interactions. FiLM and BiGated also yield competitive performance by effectively adapting modality contributions, but they still fall short of the performance of Late Fusion and Concat. Sum fusion, while simpler in its mechanism, performs less effectively, suggesting that its method of combining modality features lacks the necessary complexity to effectively capture cross-modal synergies.
\textbf{G\textsuperscript{2}D with Various Fusion Modules.}  
\Cref{tab:fusion-results} compares the effect of different fusion strategies on G\textsuperscript{2}D. Out of the traditional fusion methods used for the modality imbalance problem, \emph{late fusion} consistently achieves the best results. This highlights the effectiveness of leveraging independent unimodal representations while preserving their distinct contributions. However, \emph{Cross-Attention} based fusion closely follows the performance of late fusion, demonstrating its ability to enhance cross-modal interactions by dynamically attending to relevant features. Concat fusion provides strong performance but falls slightly behind Late Fusion. FiLM and BiGated offer adaptive feature integration yet fail to match the top-performing methods. Descriptions of these fusion strategies are in \cref{fusion-supp}.
%, while Sum fusion performs the worst, indicating that simple element-wise addition is insufficient for optimal multimodal representation learning.  

\begin{table}[t]
\centering
\normalsize
\caption{Partial vs. Complete Modality Suppression in G\textsuperscript{2}D}
\label{tab:partial-vs-complete}
\renewcommand{\arraystretch}{1.0}  % Reduces row height
\setlength{\tabcolsep}{3pt}       % Reduces column padding
\resizebox{0.9\columnwidth}{!}{
\begin{tabular}{c|cccc} 
\toprule
\textbf{Type} & \textbf{CREMA-D}        & \textbf{AV-MNIST}       & \textbf{VGGSound}       & \textbf{UR-Funny}        \\ 
\cmidrule(lr){1-5}
Partial       & 81.99          & 72.83          & 51.16          & 63.68           \\
Complete      & \textbf{85.89} & \textbf{73.03} & \textbf{53.82} & \textbf{65.49}  \\
\bottomrule
\end{tabular}
}
\end{table}

% compare G\textsuperscript{2}D's performance with partial suppression versus complete suppression strategies.
% \textbf{Modality Suppression in G\textsuperscript{2}D.}  
% We employ the activation function used by OGM-GE \cite{ogm-ge} for partial gradient suppression, while complete suppression utilizes SMP to identify dominant modalities and zero out their gradients. Results in \Cref{tab:partial-vs-complete} show that complete suppression in G\textsuperscript{2}D outperforms partial suppression across all datasets. This improvement stems from SMP’s ability to shift focus to the suppressed modality, enabling it to train to convergence without interference. As a result, complete suppression more effectively mitigates modality imbalance and enhances multimodal learning.

\textbf{Modality Suppression in G\textsuperscript{2}D.}  
We compare two suppression strategies: partial suppression and complete suppression. Partial suppression follows OGM-GE \cite{ogm-ge}, applying gradient modulation with $1-\tanh(x)$, where $x$ is the ratio of modality scores. Complete suppression utilizes SMP, which zeroes out gradients of dominant modalities, allowing the suppressed modality to train to convergence. \Cref{tab:partial-vs-complete} shows that complete suppression consistently outperforms partial suppression across all datasets. This improvement arises from SMP’s ability to shift optimization focus toward the suppressed modality, preventing interference and enabling more effective multimodal feature integration. So, complete suppression significantly mitigates modality imbalance and enhances multimodal learning.

\begin{table}[h]
\centering
\large
\caption{Effect of $\tau_{j}$ on G\textsuperscript{2}D with two \& three modalities}
\vspace{-6pt} % Reduce vertical space between text
\label{tab:tau1}
\setlength{\tabcolsep}{6pt} % Sets horizontal padding between column content to 3 points (default is 6pt)
\renewcommand{\arraystretch}{0.80} % Reduces row height spacing to 85\% of the default
\resizebox{0.80\columnwidth}{!}{
\begin{tabular}{c|cccc} 
\toprule
\textbf{$(\tau_{1},\tau_{2})$} & \textbf{(0,150)} & \textbf{(50,150)} & \textbf{(100,150)} & \textbf{(150,150)}  \\ 
\cmidrule(r){1-5}
CREMA-D                                & 78.63            & 82.80             & \uline{83.74}              & \textbf{85.89}      \\
\bottomrule
\end{tabular}
}
\end{table}
\vspace{-10pt}
\begin{table}[h]
\centering
% \caption{Effect of $\tau_{j}$ on G\textsuperscript{2}D for a dataset with three modalities}
\vspace{-6pt} % Reduce vertical space between text
\label{tab:tau2}
\setlength{\tabcolsep}{6pt} % Sets horizontal padding between column content to 3 points (default is 6pt)
\renewcommand{\arraystretch}{0.80} % Reduces row height spacing to 85\% of the default
\resizebox{0.80\columnwidth}{!}{
\begin{tabular}{c|ccc} 
\toprule
\textbf{$(\tau_{1},\tau_{2},\tau_{3})$} & \textbf{(0,0,150)} & \textbf{(50,50,150)} & \textbf{(75,75,150)}  \\ 
\cmidrule(){1-4}
IEMOCAP                                & 75.30              & \uline{76.99}        & \textbf{77.19}        \\
\bottomrule
\end{tabular}
}
\end{table}

\textbf{Effect of Prioritization Epochs ($\boldsymbol{\tau_{j}}$).} We analyze the impact of $\tau_j$, the hyperparameter controlling the epochs for each SMP stage, in \Cref{tab:tau1}. For instance, with two modalities ($k=2$), the schedule $(\tau_1, \tau_2)$ denotes epochs for training the weakest modality alone, followed by a joint training phase for both. Similarly, for three modalities ($k=3$), $(\tau_1, \tau_2, \tau_3)$ defines epochs for the weakest alone, then the second weakest alone, and finally a joint phase for all three. The results in \Cref{tab:tau1} show that systematically increasing the dedicated epochs for weaker modalities consistently improves G\textsuperscript{2}D's performance. This finding strongly validates our Hypothesis 1 (\cref{hypothesis}) that providing weaker modalities with dedicated, interference-free training phases is crucial for mitigating modality imbalance.

Additional ablations are provided in \cref{ablation-supp}.

\section{Conclusion}
In this paper, we presented G\textsuperscript{2}D, a simple but effective novel framework designed to tackle modality imbalance in multimodal learning through gradient-guided distillation and sequential modality prioritization. By fusing unimodal and multimodal learning objectives together with knowledge distillation and utilizing confidence scores from unimodal teachers, G\textsuperscript{2}D dynamically prioritizes weaker modalities. With these contributions, G\textsuperscript{2}D ensures each modality contributes effectively during training without being overshadowed by any dominant modalities. Our experimental results, which span multiple classification datasets and a regression task, illustrate that G\textsuperscript{2}D enhances feature alignment, mitigates modality imbalance, and outperforms existing state-of-the-art methods. We believe that the proposed gradient modulation strategy holds great potential to advance balanced learning in complex multimodal scenarios, paving the way for more inclusive and robust AI systems.  

\newpage

\section*{Acknowledgments}
This work is funded by the U.S. Department of Agriculture, National Institute of Food and Agriculture, Data Science for Food and Agricultural Systems (DSFAS) program. Accession number 1030694.

{
    \small
    \bibliographystyle{ieeenat_fullname}
    \bibliography{main}
}

\clearpage
\setcounter{page}{1}
\maketitlesupplementary

\appendix

\section{Appendix}

\subsection{Detailed Dataset Description}

\subsubsection{Crowd-sourced Emotional Multimodal Actors Dataset (CREMA-D) \cite{cremad}}
CREMA-D is a multimodal dataset designed for emotion recognition research. It contains audio-visual recordings of actors portraying a variety of emotional states, including Anger, Disgust, Fear, Happy, Neutral, and Sad. The dataset features actors from diverse racial and ethnic backgrounds, covering a wide age range, which makes it suitable for studying the interplay between audio and visual emotional expressions. Ratings for emotional intensity and accuracy were gathered from crowd-sourced participants. The dataset is divided into a training set of 6,027 samples, a validation set of 669 samples, and a test set of 745 samples, facilitating robust model training and evaluation.

\subsubsection{Audio Visual MNIST (AV-MNIST) \cite{avmnist}}
AV-MNIST is a synthetic multimodal dataset designed for audio-visual digit classification. It combines visual MNIST digit images, downsampled using PCA to retain 25\% of their original energy, with audio samples of spoken digits from the TIDigits dataset \cite{tidigits}. The audio samples are represented as 112 $\times$ 112 spectrograms and are augmented with noise from the ESC-50 dataset \cite{esc50}. The dataset consists of 70,000 audio-visual pairs, including 55,000 for training, 10,000 for testing, and 5,000 selected from the training set for validation.

\subsubsection{VGGSound \cite{vggsound}}
VGGSound is a large-scale audio-visual dataset designed for training and evaluating audio recognition models. It consists of over 200,000 video clips sourced from YouTube, each containing audio-visual correspondence where the sound source is visually present in the video. The dataset includes 310 diverse classes covering various real-world environments, such as people, animals, music, and nature. Each clip is 10 seconds long, ensuring both the audio and visual elements are aligned, making it ideal for audio-visual learning tasks.

\subsubsection{UR-Funny \cite{urfunny}}
UR-FUNNY is a multimodal dataset created for the task of humor detection, utilizing text, visual gestures, and prosodic acoustic cues. The dataset comprises 1,866 video clips collected from TED Talks featuring diverse speakers and covering 417 different topics. Each clip is labeled with binary humor annotations, with an equal number of humorous and non-humorous samples, ensuring a balanced dataset. The multimodal nature of UR-FUNNY makes it particularly suitable for investigating the relationships among different modalities, offering insights into how text, vision, and audio can jointly contribute to understanding humor in a multimodal learning context.

\subsubsection{IEMOCAP \cite{iemocap}}
The Interactive Emotional Dyadic Motion Capture (IEMOCAP) dataset is a widely used resource for emotion recognition, containing approximately 12 hours of audio-visual data from ten actors engaged in scripted and improvised dyadic conversations to elicit a range of authentic emotions. The dataset is rich in modalities, providing synchronized video, speech, facial motion capture data, and text transcriptions. Each utterance is annotated by multiple raters for both categorical emotions and dimensional attributes (valence, arousal, and dominance), making it suitable for diverse and nuanced modeling tasks. Due to its naturalistic dyadic interactions, IEMOCAP serves as a benchmark for developing emotion-aware conversational AI and understanding complex multimodal emotional cues.

\subsection{Details on Baselines}\label{baseline-supp}

\subsubsection{Modality-Specific Early Stopping (MSES) \cite{mses}}
MSES is a multimodal learning approach that aims to prevent overfitting by independently managing the learning rate for each modality. Within the MSES framework, each modality’s classifier is treated as a separate task, and early stopping is employed when a modality begins to overfit, while others continue to learn. By formulating a multitask setup, MSES allows for the independent regulation of modality-specific learning progress, ensuring that the stronger modalities do not overshadow the weaker ones during joint training. This method effectively prevents overfitting by identifying and stopping learning for modalities when their validation loss fails to improve, thereby maximizing balanced contributions from each modality and enhancing the overall multimodal learning process.

\subsubsection{Modality-Specific Learning Rate (MSLR) \cite{mslr}}
MSLR aims to optimize multimodal late-fusion models by assigning unique learning rates to each modality instead of using a single global learning rate. This approach helps prevent the vanishing gradients issue that can occur when learning rates are not tailored to the specific characteristics of each modality. By assigning modality-specific learning rates, MSLR ensures that each modality contributes effectively to the learning process, ultimately improving the overall performance of the multimodal model.

\subsubsection{Adaptive Gradient Modulation (AGM) \cite{agm}}
AGM addresses modality competition in multimodal models by modulating the participation level of each modality during training. Inspired by Shapley value-based attribution and the OGM-GE algorithm, AGM isolates the contribution of individual modalities and modulates their gradient update intensity accordingly, allowing stronger modalities to be suppressed while amplifying weaker ones. This adaptive strategy applies to all types of fusion architectures, thereby boosting the overall model performance by ensuring a balanced contribution from each modality and mitigating dominance effects that lead to suboptimal joint training outcomes.

\subsubsection{Prototypical Modality Rebalance (PMR) \cite{pmr}}
PMR tackles the "modality imbalance" issue by applying different learning strategies to each modality to ensure more balanced learning. Specifically, PMR uses prototypical cross-entropy (PCE) loss to accelerate the slow-learning modality, allowing it to align more closely with prototypical representations, while also reducing the inhibition from dominant modalities via prototypical entropy regularization (PER). The method effectively exploits features of each modality independently and helps prevent one modality from dominating the learning process, thereby enhancing overall multimodal learning performance.

\subsubsection{On-the-fly Gradient Modulation with Generalization Enhancement (OGM-GE) \cite{ogm-ge}}
OGM-GE addresses the issue of under-optimization for specific modalities in multimodal learning by dynamically modulating gradient contributions for each modality. This approach balances the learning pace by modulating gradients of modality-specific coefficients during backpropagation, reducing the dominance of stronger modalities and facilitating better feature exploitation of weaker ones. Additionally, OGM-GE incorporates a generalization enhancement mechanism, adding dynamic Gaussian noise to improve model generalization.

\subsubsection{Multimodal Learning with Alternating Unimodal Adaptation (MLA) \cite{mla}}
MLA addresses the issue of modality dominance by alternating the training focus between modalities rather than using conventional joint optimization. This alternating unimodal adaptation helps avoid interference between modalities, allowing each to reach its full potential while still maintaining cross-modal interactions through a shared head. A gradient modification mechanism is introduced to mitigate "modality forgetting," thereby preserving cross-modal information learned during previous iterations. At inference, MLA integrates multimodal information dynamically, using uncertainty-based fusion to manage imbalance across modality-specific contributions effectively.

\subsubsection{MMPareto: Boosting Multimodal Learning with Innocent Unimodal Assistance \cite{mm-pareto}}
MMPareto aims to enhance multimodal learning by addressing the gradient conflict that arises between unimodal and multimodal learning objectives. The algorithm uses Pareto integration to align gradient directions across objectives, ensuring a final gradient that benefits all modalities without compromising any. By balancing gradient direction and boosting gradient magnitude, MMPareto improves generalization, providing "innocent unimodal assistance" to enhance the performance of each modality while maintaining the consistency of multimodal learning.

\subsubsection{On Uni-Modal Feature Learning in Supervised Multi-Modal Learning (UMT) \cite{umt}}
This paper addresses the problem of insufficient learning of unimodal features in multi-modal learning. The proposed framework consists of two approaches: Uni-Modal Teacher (UMT) and Uni-Modal Ensemble (UME). UMT distills unimodal pre-trained features into a multi-modal model during late-fusion training, ensuring that the representations learned for each modality are preserved effectively while maintaining cross-modal interactions. UME, on the other hand, avoids cross-modal interactions by combining the predictions from unimodal models directly, thus preventing negative interference. To decide which approach to use, they employ an empirical trick explained in the paper. In our experiments, we compare against UMT due to its use of knowledge distillation (KD), which aligns with our proposed approach.

\subsubsection{ReconBoost: Boosting Can Achieve Modality Reconcilement \cite{reconboost}}
ReconBoost introduces a modality-alternating learning paradigm to mitigate modality competition in multimodal learning. Instead of optimizing all modalities simultaneously, ReconBoost updates each modality separately, ensuring that weaker modalities are not overshadowed by stronger ones. A KL-divergence-based reconcilement regularization is incorporated to maximize diversity between current and past updates, aligning the method with gradient-boosting principles. Unlike traditional boosting, ReconBoost only retains the most recent learner per modality, preventing overfitting and excessive reliance on strong modalities. Additionally, it integrates a memory consolidation regularization to preserve historical modality-specific information and a global rectification scheme to refine joint optimization. Empirical results across multiple benchmarks demonstrate that ReconBoost effectively reconciles modality learning dynamics, leading to improved multimodal fusion performance.

\subsubsection{Facilitating Multimodal Classification via Dynamically Learning Modality Gap (DLMG) \cite{dlmg}}  
DLMG addresses the modality imbalance problem in multimodal learning by focusing on disparities in category label fitting across different modalities. Unlike prior methods that primarily regulate learning rates or gradient contributions, DLMG leverages contrastive learning to align modality representations and reduce dominance effects. The approach dynamically integrates supervised classification loss and contrastive modality matching loss through either a heuristic strategy or a learning-based optimization strategy that adjusts their relative importance during training. By progressively refining modality alignment while maintaining label supervision, DLMG minimizes performance gaps between dominant and non-dominant modalities, leading to a more balanced and effective multimodal learning process.

\subsubsection{Detached and Interactive Multimodal Learning (DI-MML) \cite{di-mml}}
DI-MML proposes that modality competition is a direct result of the uniform learning objective used in traditional joint training frameworks. To eliminate this competition, DI-MML proposes a detached learning framework where each modality's encoder is trained separately with its own isolated learning objective. To enable cross-modal interaction without reintroducing competition, the framework employs two key strategies: (1) a shared classifier is used to align features from different modalities into a common embedding space, and (2) a novel Dimension-decoupled Unidirectional Contrastive (DUC) loss is introduced. The DUC loss identifies ``effective" and ``ineffective" feature dimensions within each modality and then transfers knowledge unidirectionally from the effective dimensions of one modality to the corresponding ineffective dimensions of another. This strategy facilitates the exchange of complementary information while preserving the integrity of each modality's well-learned features.

\subsection{Details on Fusion Techniques}\label{fusion-supp}

\subsubsection{Summation}
Summation fusion is a straightforward multimodal integration technique where features from multiple modalities are combined through element-wise addition. Each modality contributes independently, and their respective representations are directly summed without any complex cross-modal interactions. In this approach, the output of each modality-specific encoder is first processed by a fully connected layer to generate unimodal predictions, which are then added together to form a unified representation. This combined output is used to compute a loss, which subsequently updates all components involved, including the encoders and the fully connected layers. Summation fusion's strength lies in its simplicity and ease of implementation, as it does not require intricate fusion mechanisms. However, it does not explicitly capture inter-modal relationships, potentially limiting its effectiveness in scenarios where richer cross-modal interactions are beneficial.

\subsubsection{Concatenation}
Concatenation fusion is a common strategy for integrating information from different modalities by concatenating their feature vectors along a specified axis. This method combines feature representations directly, allowing the model to consider information from all modalities together as a single, extended vector. Despite enabling joint perception of multimodal data, it does not explicitly model cross-modal interactions. The concatenated features are passed through a fully connected layer, where the input size equals the sum of all encoder output dimensions, and the output size matches the number of classes. During training, the model uses the resulting fusion output to compute the loss and update all the involved parameters, including those of the individual encoders and the fully connected layer. Concatenation fusion is effective in creating a unified feature representation, but it relies on subsequent layers to extract and learn any interactions between the modalities.

\subsubsection{Feature-wise Linear Modulation (FiLM) \cite{film}}
FiLM is a sophisticated fusion method that integrates information from multiple modalities by adjusting feature representations in one modality according to the information from another. This modulation approach uses conditional inputs to produce parameters that scale and shift feature activations, enabling the model to dynamically adjust its processing based on context. FiLM works by passing the conditioning modality through a layer that outputs these modulation parameters, which then directly adjust the target modality's features before they proceed to the next layers in the model. By providing targeted feature modulation, FiLM helps capture cross-modal nuances and allows the model to be more adaptive in multimodal learning tasks that require context-sensitive adjustments.

\subsubsection{BiLinear Gated Fusion (BiGated) \cite{gated}}
BiGated fusion combines bilinear pooling and gating mechanisms to enhance the integration of multiple modalities by capturing their complex interactions. This technique explicitly models cross-modal relationships, providing a more expressive and fine-grained fusion strategy compared to simpler approaches like concatenation or summation. In BiGated fusion, each modality passes through its own fully connected layer, much like summation. However, what sets BiGated apart is its use of a gating mechanism—one modality's hidden state is processed through an activation function (we use sigmoid) to generate a gated weight, which is then used to modulate the contributions of other modalities. This approach ensures that each modality can dynamically influence how other modalities are represented in the fusion process, allowing for a richer and more adaptive multimodal representation before proceeding to the final classification layers.

\subsubsection{Cross-Attention Fusion \cite{cross-attention}}  
Cross-attention fusion enables the dynamic and adaptive integration of multiple modalities by allowing each modality to attend to others through bidirectional or all-directional attention mechanisms. This approach explicitly models inter-modal dependencies, ensuring that each modality can selectively focus on the most relevant features from others. In our implementation, for two-modal cases, modality $X$ attends to modality $Y$ and vice versa, refining their representations based on mutual interactions. For three-modal scenarios, all-directional attention is applied, where each modality interacts not only with one other but also with the third, ensuring comprehensive multimodal integration. The attended representations are normalized to enhance stability and mitigate potential imbalances in feature contributions. Finally, the refined features from all modalities are projected into a unified representation through a fully connected layer. This mechanism effectively captures nuanced cross-modal relationships, allowing the model to leverage complementary modality-specific information for robust multimodal learning.

\subsubsection{Late Fusion \cite{late-fusion}}
Late fusion technique involves independently processing each modality through its respective model or encoder, followed by combining the outputs at a later stage to produce the final prediction. This approach allows each modality to be modeled and optimized in isolation, maintaining the unique properties of each data source. However, it may miss opportunities to exploit early cross-modal interactions that could provide additional benefits during feature learning. In late fusion, each modality-specific encoder is followed by its own fully connected layer, which is trained solely on that modality's data. The fusion output is computed by averaging the outputs of all unimodal models, ensuring that the fusion occurs only after independent learning is complete. This independence provides flexibility and robustness, especially in scenarios where some modalities may be missing, but limits the ability to deeply integrate multimodal relationships early in the learning process.

\subsection{Details on Experimental Setups}\label{experimental-setup-supp}

\subsubsection{Model Architectures}

\paragraph{ResNet-18} 
ResNet-18, a convolutional neural network with 18 layers, belongs to the ResNet family and is renowned for addressing the vanishing gradient problem through residual connections. These residual connections allow information to bypass some layers, which helps stabilize training even in deeper networks. In our experiments, we use ResNet-18 as an encoder for both audio and video modalities across CREMA-D, AV-MNIST, and VGGSound datasets. We used a specific weight initialization strategy: Xavier normal for fully connected layers, Kaiming normal for convolutional layers, and constant initialization for batch normalization layers, which facilitated an effective starting point for network training and ensured stable convergence across multimodal tasks.

\paragraph{Transformer}
Transformers are powerful architectures designed for handling sequential data and capturing long-range dependencies through self-attention mechanisms. In our implementation, we employ Transformers as encoders for the audio, video, and text modalities of the UR-Funny and IEMOCAP dataset. Following the approach described in \cite{agm}, we utilized a 4-layer Transformer encoder with eight attention heads and a hidden dimension of 768 for each modality in the UR-Funny and IEMOCAP dataset. Input features were projected to a 768-dimensional embedding using a convolutional layer, ensuring consistency across different modalities. We employed a similar initialization strategy to ResNet-18 to facilitate stable training.

\subsubsection{Hyperparameters} \label{hyperparameter-supp}
We trained our models on 1 Nvidia A10 GPU with a batch size of 16 using the SGD optimizer, with a momentum of 0.9 and a weight decay of $1 \times 10^{-4}$. We initialized the learning rate at 0.001 and decayed it by a ratio of 0.1 every 200 epochs. For all experiments, we set the random seed to 999 for reproducibility. We defined the G\textsuperscript{2}D loss function as a weighted sum of student loss, feature loss, and logit loss, where $\alpha$ and $\beta$ are weighting coefficients for the feature loss and logit loss, respectively. We set both $\alpha$ and $\beta$ to 1.0 for all datasets. Additionally, for the logit loss, we used a temperature of 1.0 in the KL Divergence without further softening, effectively utilizing hard logits for the training process.

\begin{table}[b]
\centering
\large
\caption{Comparing G\textsuperscript{2}D with DI-MML}
\label{tab:di-mml}
\setlength{\tabcolsep}{6pt} % Sets horizontal padding between column content to 3 points (default is 6pt)
\renewcommand{\arraystretch}{1.0} % Reduces row height spacing to 85\% of the default
\resizebox{0.7\columnwidth}{!}{
\begin{tabular}{c|ccc} 
\toprule
Method   & Joint-Train & \textcolor[rgb]{0.2,0.2,0.2}{~DI-MML} & G\textsuperscript{2}D  \\ 
\cmidrule(r){1-4}
CREMA-D  & 67.47       & \uline{83.51}                                 & \textbf{85.89}         \\
AV-MNIST & 69.77       & \uline{71.35}                                 & \textbf{73.03}         \\
\bottomrule
\end{tabular}
}
\end{table}

\subsection{Comparison of G\textsuperscript{2}D with DI-MML}\label{di-mml-comparison}
In response to reviewer feedback, we provide an additional comparison against the state-of-the-art baseline DI-MML~\cite{di-mml}. As shown in Table~\ref{tab:di-mml}, G\textsuperscript{2}D outperforms DI-MML on both the CREMA-D and AV-MNIST datasets. This result further validates the effectiveness of G\textsuperscript{2}D in relation to current leading methods in the field.

\subsection{Further Analysis of G\textsuperscript{2}D}\label{g2d-analysis-supp}

\subsubsection{Distinction of G\textsuperscript{2}D from UMT and OGM-GE}
The primary novelty of G\textsuperscript{2}D arises from its unique \textit{G\textsuperscript{2}D loss} and its Sequential Modality Prioritization (SMP) technique, and critically, from their synergistic combination. 

Our \textit{G\textsuperscript{2}D loss} improves upon the distillation strategy of UMT \cite{umt} by incorporating a KL divergence-based logit loss. This addition is crucial for enabling the student model to learn the nuanced inter- and intra-class relationships captured by the unimodal teachers' soft logits.

Furthermore, our SMP technique is fundamentally different from the gradient modulation in OGM-GE \cite{ogm-ge} in two principal ways:
\begin{enumerate}
    \item \textbf{Guidance for Modulation:} OGM-GE calculates modality confidence from the student's own encoders during training. This signal can be noisy and unreliable, especially in early stages. In contrast, SMP leverages stable confidence scores from \textit{pre-trained unimodal teachers}, providing a more robust and accurate signal to identify weaker modalities automatically.
    \item \textbf{Suppression Mechanism:} OGM-GE uses functions like $1-\tanh(\cdot)$ to only \textit{partially} suppress dominant modalities, meaning they continue to train simultaneously and modality competition can persist. SMP enforces a \textit{complete gradient shutdown} for non-prioritized modalities. This ensures that the prioritized weak modality trains in true isolation, more effectively mitigating interference from dominant modalities.
\end{enumerate}
The results in \Cref{tab:components} validate these distinctions, showing that the \textit{G\textsuperscript{2}D loss} alone surpasses UMT, SMP alone surpasses OGM-GE, and their combination yields the best overall performance.

\begin{table}[t]
\centering
\large
\caption{Comparing Components of G\textsuperscript{2}D with UMT \& OGM-GE}
\label{tab:components}
\setlength{\tabcolsep}{4pt} % Sets horizontal padding between column content to 3 points (default is 6pt)
\renewcommand{\arraystretch}{1.0} % Reduces row height spacing to 85\% of the default
\resizebox{1.0\columnwidth}{!}{
\begin{tabular}{c|c|cc|cc|c} 
\toprule
\textbf{Method} & \begin{tabular}[c]{@{}c@{}}\textbf{Joint-}\\\textbf{Train}\end{tabular} & \textbf{UMT} & \begin{tabular}[c]{@{}c@{}}\textbf{G\textsuperscript{2}D}\\\textbf{Loss}\end{tabular} & \begin{tabular}[c]{@{}c@{}}\textbf{OGM-}\\\textbf{GE}\end{tabular} & \textbf{SMP} & \begin{tabular}[c]{@{}c@{}}\textbf{G\textsuperscript{2}D}\\\textbf{(SMP +~\textbf{G\textsuperscript{2}D Loss })}\end{tabular}  \\ 
\cmidrule(lr){1-7}
CREMA-D         & 67.47                                                                   & 67.61        & 78.63                                                                                 & 72.18                                                              & \uline{80.78}        & \textbf{85.89}                                                                                                                 \\
AV-MNIST        & 69.77                                                                   & 72.33        & \uline{72.76}                                                                                 & 71.08                                                              & 72.51        & \textbf{73.03}                                                                                                                 \\
\bottomrule
\end{tabular}
}
\end{table}

\subsubsection{Synergy of Distillation and Sequential Modality Prioritization}
The motivation for integrating our \textit{G\textsuperscript{2}D loss} (via KD) with SMP is to address the limitations of using either technique alone. The distillation component leverages unimodal teachers—trained in isolation—to provide the student with stable, competition-free feature and logit targets. This guides the student towards more balanced representations than learning solely from GT labels amidst modality competition. 

However, even with this guidance, the student's modality encoders are still optimized simultaneously, which allows modality imbalance to persist. SMP is introduced to solve this. The crucial \emph{synergy} lies in the fact that during the isolated training phases enforced by SMP, the prioritized weak modality learns not only from the ground-truth labels but also from the rich, interference-free knowledge distilled from its unimodal teacher via the \textit{G\textsuperscript{2}D loss}. This focused, dual-signal learning in an isolated context enables the robust development of weaker modalities. By combining SMP with our distillation objective, G\textsuperscript{2}D mitigates modality imbalance more thoroughly and effectively than using either technique independently.

\begin{table}[t]
\centering
\large
\caption{Single-Batch Resource Metrics on CREMAD}
\label{tab:overhead}
\setlength{\tabcolsep}{6pt} % Sets horizontal padding between column content to 3 points (default is 6pt)
\renewcommand{\arraystretch}{1.0} % Reduces row height spacing to 85\% of the default
\resizebox{\columnwidth}{!}{
\begin{tabular}{c!{\vrule width \lightrulewidth}c!{\vrule width \lightrulewidth}c} 
\toprule
\textbf{Method}       & \textbf{Total Memory (MB)} & \textbf{Total Execution Time (ms)}  \\ 
\midrule
Joint-Train           & 12.1366 MB                 & 4531.8                              \\
G\textsuperscript{2}D & 12.1998 MB                 & 4539.8                              \\
\bottomrule
\end{tabular}
}
\end{table}

\subsubsection{Computational Cost}
The training overhead of G\textsuperscript{2}D is negligible, and there is no additional overhead during inference. This efficiency stems from the framework's design: unimodal teacher models are pre-trained, and their outputs (e.g., logits and features) are saved. During the multimodal student model's training, these pre-computed outputs are loaded from disk per batch in a process analogous to loading the dataset itself. As quantified in \Cref{tab:overhead}, for a typical 16-sample batch, G\textsuperscript{2}D requires only $\approx$0.5\% more memory and adds merely $\approx$0.15\% to the execution time compared to a standard joint-training baseline. Therefore, in resource-constrained environments, if a traditional joint-training model is feasible, G\textsuperscript{2}D is also readily viable by performing the one-time teacher training first and then training the student.

\subsection{Additional Ablation Studies}\label{ablation-supp}

\begin{table}[b]
\centering
\Large
\caption{G\textsuperscript{2}D vs. Missing Modality Methods on IEMOCAP}
\label{tab:missing-modality}
\setlength{\tabcolsep}{2pt} % Sets horizontal padding between column content to 3 points (default is 6pt)
\renewcommand{\arraystretch}{1.0} % Reduces row height spacing to 85\% of the default
\resizebox{1.0\columnwidth}{!}{
\begin{tabular}{c|c|cccc|c} 
\toprule
\textbf{Miss Rate} & \textbf{Joint-Train} & \textbf{CRA \cite{cra}}  & \textbf{MMIN \cite{mmin}} & \textbf{CPM-Net \cite{cpm-net}} & \textbf{TATE \cite{tate}} & \textbf{G\textsuperscript{2}D}  \\ 
\cmidrule(lr){1-7}
0\%                & 75.51                & \uline{76.21} & 74.94         & 58.00            & 69.92         & \textbf{77.19}                  \\
20\%               & 69.06                & 67.34         & \uline{69.36} & 53.65            & 63.22         & \textbf{71.49}                  \\
40\%               & 61.09                & 57.04         & \uline{63.30} & 51.01            & 60.36         & \textbf{65.10}                  \\
60\%               & 52.41                & 43.22         & 57.52         & 47.38            & \uline{57.99} & \textbf{61.50}                  \\
\bottomrule
\end{tabular}
}
\end{table}

\subsubsection{Learning with Missing Modalities}
To evaluate the robustness of G\textsuperscript{2}D with incomplete data, we conduct experiments on the IEMOCAP dataset with randomly missing modalities, generating miss rate masks from 0\% to 60\% following the setup in MLA~\cite{mla}. As demonstrated in \Cref{tab:missing-modality}, G\textsuperscript{2}D consistently outperforms several state-of-the-art methods designed specifically for this task across all tested rates. This superior performance, even as data becomes highly sparse, suggests that by mitigating modality imbalance and fostering well-rounded representations, our framework learns more resilient features that are less dependent on any single data source, making it inherently more robust when modalities are unavailable.

\begin{table}[h]
\centering
\large
\caption{Effect of $\alpha$ and $\beta$ in G\textsuperscript{2}D with SMP}
\label{tab:alpha-beta_results}
\setlength{\tabcolsep}{3pt}       % Reduces column padding
\resizebox{\columnwidth}{!}{
\begin{tabular}{c|c|c|c|c|c|c|c} 
\toprule
\multirow{2}{*}{\textbf{Dataset }} & \multicolumn{7}{c}{($\mathbf{\alpha}$, $\mathbf{\beta}$) \textbf{weights}}                                      \\ 
\cline{2-8}
                                   & (0, 0) & (0.25, 0.75)  & (0.5, 0.5)    & (0.75, 0.25) & (1, 0) & (0, 1) & (1, 1)          \\ 
\hline
CREMA-D                            & 80.78  & 84.41         & \uline{84.95} & 84.81        & 82.39  & 84.68  & \textbf{85.89}  \\
UR-Funny                           & 63.58  & \uline{64.79} & 64.29         & 64.29        & 64.59  & 64.69  & \textbf{65.49}  \\
\bottomrule
\end{tabular}
}
\end{table}

\subsubsection{Effect of $\alpha$ and $\beta$ in G\textsuperscript{2}D.} $\alpha$ and $\beta$ denote the weighting coefficients of feature loss and logit loss, respectively in the proposed G\textsuperscript{2}D loss. Table~\ref{tab:alpha-beta_results} evaluates the effect of changing the weightage of feature loss and logit loss on G\textsuperscript{2}D. Assigning full weight to both losses ($\alpha = 1, \beta = 1$) yields the best overall performance, highlighting their combined importance in multimodal learning. In contrast, removing both losses ($\alpha = 0, \beta = 0$) significantly reduces performance, confirming their necessity. While different weight combinations impact results, incorporating both losses with higher weight leads to greater improvements across datasets. 

\end{document}